%% file: acl_latex.tex
\PassOptionsToPackage{table}{xcolor}
\documentclass[11pt]{article}

\usepackage[final]{acl}

\usepackage{times}
\usepackage{latexsym}
\usepackage{xcolor}            
\usepackage[most]{tcolorbox}
\usepackage{listings}          
\usepackage{makecell}
\usepackage{titlesec}          
\usepackage{hyperref}
\usepackage{float}
\usepackage{caption}
\definecolor{royalblue(web)}{rgb}{0.25, 0.41, 0.88}
\definecolor{blue-violet}{rgb}{0.54, 0.17, 0.89}
\definecolor{brightmaroon}{rgb}{0.76, 0.13, 0.28}
\definecolor{darkmagenta}{rgb}{0.55, 0.0, 0.55}
\definecolor{bleudefrance}{rgb}{0.19, 0.55, 0.91}
\definecolor{palatinateblue}{rgb}{0.15, 0.23, 0.89}
\definecolor{royalblue(web)}{rgb}{0.25, 0.41, 0.88}
\definecolor{whitesmoke}{rgb}{0.96, 0.96, 0.96}
\definecolor{thulianpink}{rgb}{0.87, 0.44, 0.63}
\definecolor{amber(sae/ece)}{rgb}{1.0, 0.49, 0.0}
\definecolor{darkblue}{rgb}{0.0, 0.0, 0.55}
\definecolor{alizarin}{rgb}{0.82, 0.1, 0.26}
\definecolor{asparagus}{rgb}{0.53, 0.66, 0.42}
\definecolor{darkspringgreen}{rgb}{0.09, 0.45, 0.27}
\definecolor{columbiablue}{rgb}{0.61, 0.87, 1.0}
\definecolor{wildblueyonder}{rgb}{0.64, 0.68, 0.82}
\definecolor{trolleygrey}{rgb}{0.5, 0.5, 0.5}
\definecolor{paleaqua}{rgb}{0.74, 0.83, 0.9}
\definecolor{bubblegum}{rgb}{0.99, 0.76, 0.8}
\definecolor{coralred}{rgb}{1.0, 0.25, 0.25}
\definecolor{green(ryb)}{rgb}{0.4, 0.69, 0.2}
\definecolor{flame}{rgb}{0.89, 0.35, 0.13}
\definecolor{bittersweet}{rgb}{1.0, 0.44, 0.37}
\definecolor{darksalmon}{rgb}{0.91, 0.59, 0.48}
\definecolor{emerald}{rgb}{0.31, 0.78, 0.47}
\definecolor{green(pigment)}{rgb}{0.0, 0.65, 0.31}

\lstdefinestyle{PythonCode}{
    language=Python,
    frame=none,
    basicstyle=\ttfamily,
    breaklines=true,
    breakatwhitespace=false, 
    breakautoindent=true, 
    keywordstyle=\bfseries\color{bleudefrance},
    morekeywords={},
    emph={self},
    emphstyle=\bfseries\color{darkmagenta},
    commentstyle=\itshape\color{black!50!white},
    stringstyle=\bfseries\color{darkblue},
    columns=flexible,
    breaklines=true,
    breakatwhitespace=true
}

\definecolor{codegreen}{rgb}{0,0.6,0}
\definecolor{codegray}{rgb}{0.5,0.5,0.5}
\definecolor{codepink}{RGB}{252, 142, 172}
\definecolor{codepurple}{rgb}{0.58,0,0.82}
\definecolor{backcolour}{RGB}{245,245,245}

\lstdefinestyle{mystyle}{
    backgroundcolor=\color{backcolour},   
    commentstyle=\color{magenta},
    keywordstyle=\color{blue},
    numberstyle=\tiny\color{codegray},
    stringstyle=\color{codepurple},
    basicstyle=\fontfamily{\ttdefault}\footnotesize,
    breakatwhitespace=false,         
    breaklines=true,                 
    keepspaces=true,    
    frame=single,
    numbersep=5pt,                  
    showspaces=false,                
    showstringspaces=false,
    showtabs=false,                  
    tabsize=2,
    classoffset=1, 
    keywordstyle=\color{violet}, 
    classoffset=0,
}
\input{math_commands.tex}

\usepackage{hyperref}
\usepackage{url}
\usepackage{graphicx}
\usepackage{booktabs} 
\usepackage{multirow}
\usepackage{colortbl}
\usepackage{wrapfig}
\usepackage{subcaption}

\usepackage{tikz}
\usepackage[export]{adjustbox} 

\usepackage[T1]{fontenc}

\usepackage[utf8]{inputenc}

\usepackage{microtype}

\usepackage{inconsolata}

\usepackage{graphicx}

\title{Generative Interfaces for Language Models}

\author{
  Jiaqi Chen\thanks{\ First two authors contributed equally.}\thanks{\ Project done while visiting Stanford.}$^1$, 
  Yanzhe Zhang\footnotemark[1]$^2$, 
  Yutong Zhang$^1$, 
  Yijia Shao$^1$, 
  Diyi Yang$^1$ \\
  $^1$Stanford University
  $^2$Georgia Tech \\
  \texttt{chenjq24@stanford.edu}, \texttt{z\_yanzhe@gatech.edu}, \texttt{diyiy@stanford.edu}\\
}

\begin{document}
\maketitle
\begin{abstract}
\input{files/0-abs}
\end{abstract}
\input{files/1-intro}
\input{files/3-method}
\input{files/4-exp}
\input{files/2-related}
\input{files/5-con}
\bibliography{cited}

\appendix
\input{files/appendix}

\end{document}

%% file: math_commands.tex
\usepackage{amsmath,amsfonts,bm}

\def\eqref#1{equation~\ref{#1}}

\def\1{\bm{1}}

\DeclareMathAlphabet{\mathsfit}{\encodingdefault}{\sfdefault}{m}{sl}
\SetMathAlphabet{\mathsfit}{bold}{\encodingdefault}{\sfdefault}{bx}{n}



%% file: files/0-abs.tex
Large language models (LLMs) are increasingly seen as assistants, copilots, and consultants, capable of supporting a wide range of tasks through natural conversation. However, most systems remain constrained by a linear request-response format that often makes interactions inefficient in multi-turn, information-dense, and exploratory tasks. To address these limitations, we propose \textbf{Generative Interfaces} for Language Models, a paradigm in which LLMs respond to user queries by proactively generating user interfaces (UIs) that enable more adaptive and interactive engagement. Our framework leverages structured interface-specific representations and iterative refinements to translate user queries into task-specific UIs. For systematic evaluation, we introduce a multidimensional assessment framework that compares generative interfaces with traditional chat-based ones across diverse tasks, interaction patterns, and query types, capturing functional, interactive, and emotional aspects of user experience. Results show that generative interfaces consistently outperform conversational ones, with up to a 72\% improvement in human preference. These findings clarify when and why users favor generative interfaces, paving the way for future advancements in human-AI interaction.
Data and code are available at \url{https://github.com/SALT-NLP/GenUI}.

%% file: files/1-intro.tex
\input{figures/1-intro}
\section{Introduction}
A longstanding goal in computing is to design systems that not only respond to users but also adapt by dynamically reshaping interfaces to facilitate users' interaction and help them achieve their goals~\citep{apple1987knowledgeNavigator, lyytinen2002ubiquitous}. While recent advances in large language models (LLMs) have brought us closer to this vision by enabling flexible natural language understanding, the dominant interaction paradigm, which we call the conversational UI, remains static and linear: most LLM outputs are still rendered as long blocks of text, regardless of task complexity or user preference, limiting the model’s ability to support the diverse ways users seek to learn, explore, and interact. At the same time, state-of-the-art LLMs have shown remarkable capabilities in automatically generating high-quality, functional webpages from sketches, queries, or natural language descriptions \citep{si2024design2code, li2024sketch2code,xiao2024interaction2code}.
Together, these developments raise an exciting research question: \emph{\textbf{How can LLMs go beyond conversational interfaces to enable adaptive, goal-driven interactions that meaningfully serve human needs?}}

In this work, we introduce \textbf{Generative Interfaces}, a new paradigm that differs from conversational UIs. Rather than delivering static text responses within a predefined chatbot window, Generative Interfaces dynamically create entirely new interface structures that adapt to users' specific goals and interaction requirements.
While recent tools like OpenAI's Canvas \citep{canvas} and Claude's Artifacts \citep{artifacts} enhance user interaction by providing dedicated workspaces for documents, code, and visualizations, our approach extends this vision by supporting deeper engagement and enabling richer, task-specific experiences.
For example, as shown in Figure~\ref{fig:intro}, when users pose questions such as ``\textit{I want to understand neural networks}'' or ``\textit{How can I learn piano effectively?}'', conversational interfaces typically return long blocks of text. In contrast, Generative Interfaces transform these queries into an interactive neural network animation or a piano practice tool that offers real-time feedback.
This paradigm shift presents two key challenges: (I) building the infrastructure to generate user interfaces on the fly in response to users' queries, and (II) rigorously evaluating whether such generated interfaces actually improve user experience.

To address the first challenge, our framework introduces a \textbf{structured interface-specific representation} coupled with an \textbf{iterative refinement} procedure.
The structured representation enables more controllable and interpretable generation by explicitly modeling high-level interaction flows, interface state transitions, and component dependencies, which we formalize using finite state machines \citep{shehady1997FSMmethod, wagner2006modelingFSM}.
The iterative refinement procedure further enhances output quality by prompting LLMs to generate query-specific evaluation rubrics and repeatedly refine interface candidates through generation-evaluation cycles until the system converges on a polished, context-appropriate solution.
To address the second challenge, we establish a systematic evaluation framework for assessing language model interfaces across three key dimensions: functionality, interactivity, and emotional perception~\citep{hartmann2008towards, nielsen2012usability, duan2025systematic}. Specifically, we construct a diverse prompt suite, \textbf{\textit{U}}ser \textbf{\textit{I}}nterface e\textbf{\textit{X}}perience (UIX), that strategically covers diverse domains and prompt types to reflect real-world usage scenarios~\citep{tamkin2024clio}. For each user query, we recruit experienced annotators to interact with different interfaces and conduct pairwise comparisons.
Beyond this fixed prompt suite, we further conduct a complementary evaluation involving real users and their self-reported queries, which substantiates the advantages of generative interfaces under more open-ended and authentic usage conditions.
We also reveal \textit{when} they excel (in structured and information-dense domains) and \textit{why} users prefer them (through enhanced visual organization, interactivity, and reduced cognitive load).

Our main contributions are as follows:
(I) We propose \textbf{Generative Interfaces}, a paradigm that enables adaptive, goal-driven interactions with LLMs by dynamically generating user interfaces.
(II) We develop a technical infrastructure with structured representations and iterative refinement, and an evaluation framework that systematically compares generative and conversational interfaces.
(III) We demonstrate that generative interfaces significantly outperform conversational ones across diverse query types and interaction patterns.

%% file: figures/1-intro.tex
\begin{figure*}[h]
  \centering
  \includegraphics[width=1\linewidth]{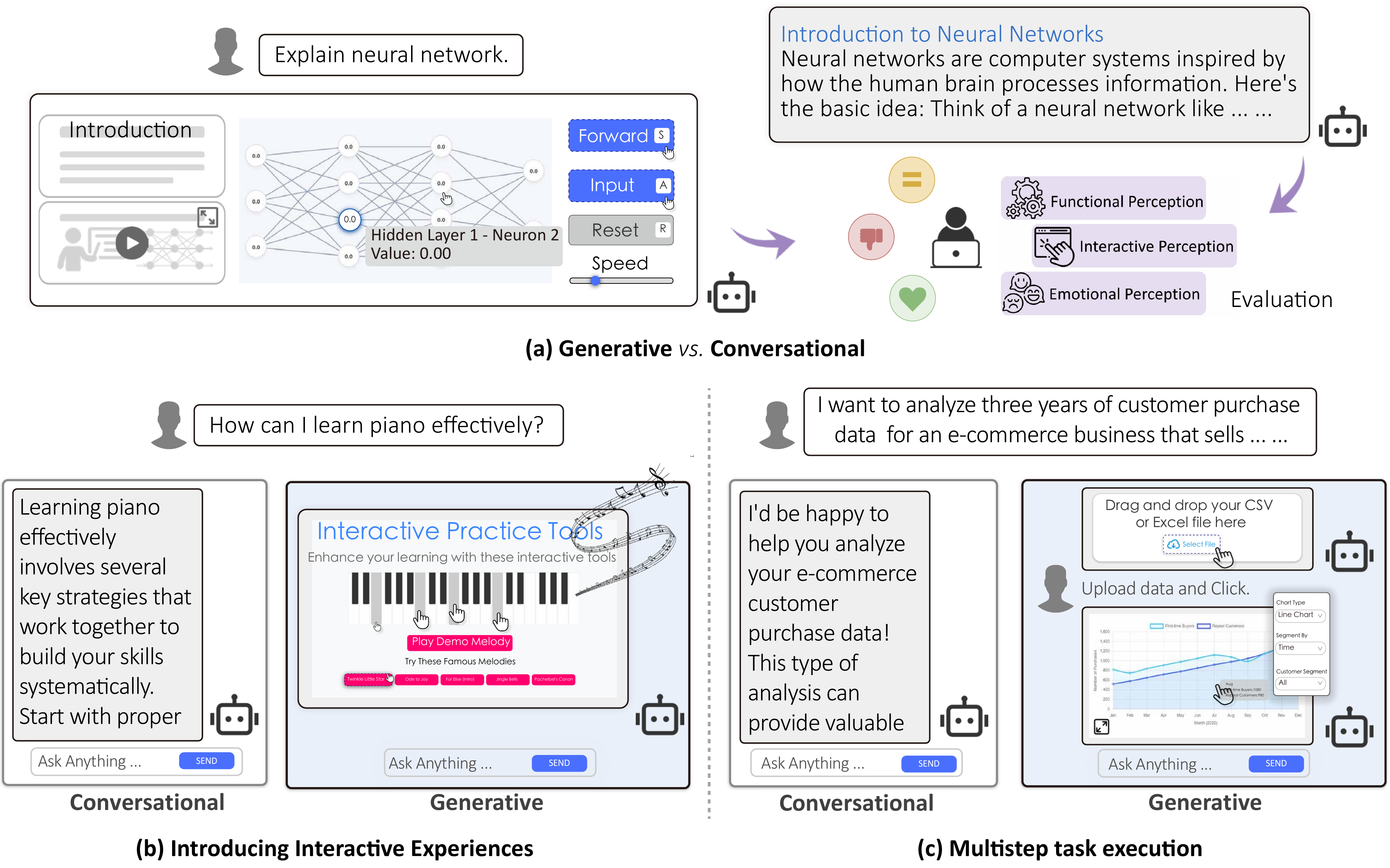}
  \caption{\textbf{Generative Interfaces compared to conversational interfaces.} 
  (a) Conceptual framework showing how Generative Interfaces create structured, interactive experiences rather than static text responses, evaluated along functional, interactive, and emotional dimensions. (b–c) Example queries illustrate how Generative Interfaces transform user input into adaptive tools—such as interactive learning aids or multistep workflows—providing clearer organization and richer interactivity than conversational responses.}
  \label{fig:intro}
\end{figure*}

%% file: files/3-method.tex
\section{Generative Interface for LMs}
\label{sec:prototype}
We introduce the structured interface-specific representation (Sec .~\ref {sec:representation}), outline the generation pipeline (Sec .~\ref {sec:inference}), and finally describe the iterative refinement using adaptive reward functions (Sec .~\ref {sec:iterative_refinement}).

\input{figures/2-method}
\subsection{Structured Interface-specific Representation}
\label{sec:representation}
Directly generating interfaces is challenging due to the vast search space and the complexity of interactive contexts.
To address this, we prompt LLMs to translate user queries into a \textbf{structured interface-specific representation} that anchors and guides the generation process.

This representation operates at two complementary levels: \textbf{(I)} \emph{high-level} interaction flows that capture user trajectories and task phases, and \textbf{(II)} \emph{low-level} finite state machines (FSMs) that define component behaviors and UI logic.

\noindent
\textbf{Interaction flows}
The high-level interaction flow provides a symbolic abstraction of user behavior across primary interface stages. It represents user task progression as a directed graph, where transitions are triggered by UI events such as clicking.
We denote this abstraction as a directed graph $\mathcal{G} = (\mathcal{V}, \mathcal{T})$, where nodes $\mathcal{V}$ represent interface views or subgoals, and edges $\mathcal{T}$ denote possible transitions (See Appendix~\ref{appendix:representation} for detail definition).
In the example shown in Figure~\ref{fig:method}, the natural language query \textit{I want to understand quantum physics principles''} is grounded into a coherent interaction trajectory: \texttt{Open Home View~$\rightarrow$Explore Tutorials$\rightarrow$Run Simulation$\rightarrow$~Glossary Lookup}''.
This abstraction captures the high-level intent and interaction logic of potential users, while concrete UI behaviors (e.g., state toggles and modal updates) are specified separately in the FSM.

\noindent
\textbf{Finite state machines} 
We further use Finite State Machines (FSMs) to describe how individual UI modules respond to user actions and update their states accordingly.
Formally, we model each UI component as
$\mathcal{M} = (\mathcal{S}, \mathcal{E}, \delta, s_0)$,
where $\mathcal{S}$ is the set of atomic interface states (\textit{e.g.}, \texttt{isModalOpen=true}), $\mathcal{E}$ is the set of user-triggered events (\textit{e.g.}, click, hover), $\delta$ is the state transition function, and $s_0$ is the initial state (See Appendix~\ref{appendix:representation}). This structure explicitly defines how the interface should behave given a particular state and a triggered event.

\subsection{Generation pipeline}
\label{sec:inference}

The whole generation pipeline is built on multiple LLM generation steps at runtime.

\noindent
\textbf{(I) Requirement specification} Starting from the user query, we first generate a requirement specification that captures the main goal, desired features, UI components, interaction styles, and problem-solving strategies. This specification serves as a bridge between the user’s natural language intent and formal interface design.

\noindent
\textbf{(II) Representation generation} Second, we generate a structured interface-specific representation (Sec.~\ref{sec:representation}) based on the requirement specification.
This representation serves as a modular and interpretable scaffold for UI generation, where the hierarchy of interaction flows and finite state machines ensures that the resulting interfaces are both coherent and functional.

\noindent
\textbf{(III) UI generation} To support the UI generation based on structured representations, we build a complementary codebase containing reusable implementations of common UI elements (\textit{e.g.}, clock, map, calculator, video player, code viewer, and chart).
Additionally, a web retrieval module\footnote{We use \url{exa.ai} as the search API.} gathers relevant UI examples and data sources.
Finally, the entire context, including the natural language query, requirement specification, structured representation, predefined components, and retrieved content examples, is passed to an LLM to synthesize executable HTML/CSS/JS code, which is rendered into an interface, as illustrated in Figure~\ref{fig:method}.

\subsection{Iterative UI Refinement}
\label{sec:iterative_refinement}

Generating an effective and well-structured user interface is usually an iterative process \citep{li2024sketch2code}. To this end, we introduce an adaptive, reward-driven iterative refinement procedure that progressively improves UI quality by generating evaluation metrics, scoring candidates, and regenerating interfaces through multiple cycles.

\noindent
\textbf{(I) Adaptive reward function}
To support task-specific and context-aware evaluation, we employ an LLM to construct a reward function tailored to each user query adaptively. As shown in Figure~\ref{fig:method}(e), for query \textit{“I want to understand quantum physics principles,”} the system automatically generates a set of fine-grained evaluation metrics—such as \textit{Visual Structure}, \textit{Explain Physics Concept}, and \textit{Clarity}—each with associated weights and verification rules. These dimensions are scored independently and aggregated to compute the final overall score, which ranges from 0 to 100. See Appendix~\ref{appendix:reward_function} for examples of adaptive reward functions.

\noindent
\textbf{(II) Iterative refinement}
As depicted in Figure~\ref{fig:method}(d), at each iteration, multiple UI candidates are generated, then the adaptive reward function evaluates these candidates.
In the next iteration, we will regenerate the UI using the highest-scoring candidate from the previous iteration, along with its evaluation.
This feedback loop guides the LLM to address issues related to structure, semantics, or visual design. The process continues until a candidate reaches an overall score of 90 or higher, or until we have completed five iterations.

\section{Evaluation Framework}
To enable systematic evaluation, we developed a comprehensive evaluation framework, which includes a diverse user query suite named \textbf{\textit{U}}ser \textbf{\textit{I}}nterface e\textbf{\textit{X}}perience (UIX), covering various scenarios, styles, and intents (Sec.~\ref{sec:prompt_suite}); a set of multidimensional evaluation metrics (Sec.~\ref{sec:evaluation_metrics}); and an integrated human study (Sec.~\ref{sec:human_study}).

\label{sec:evaluation}
\subsection{User Queries}
\label{sec:prompt_suite}
In UIX, we generate a test set of 100 user queries using Claude 3.7 that spans multiple domains, supports varying specificity levels, and captures different query complexities (See Appendix~\ref{appendix:prompt_suite} for details).
Specifically, we follow best practices from prior work around how people engage with LLMs as follows. 
\textbf{(I) Topic coverage}: Prompts are uniformly distributed across the ten domains defined in Clio \citep{tamkin2024clio}, covering a wide range of real-world user scenarios.
\textbf{(II) Query detail level}: Following \citet{Cao_2025}, each domain contains an equal split of concise and detailed prompts.
Concise prompts express intent abstractly in fewer than 15 words (\textit{e.g.}, “\textit{Create a SWOT analysis for my small business}”), while detailed prompts provide explicit goals and rich context.
\textbf{(III) Query Type}: As user queries shift from casual dialogue to actionable tasks, our design maintains a balanced mixture between general conversational prompts (\textit{e.g.}, ``\textit{How can I improve my public speaking?}”) and interactive, task-oriented queries (\textit{e.g.}, ``\textit{I want to visualize my company's sales data}”).

\subsection{Evaluation Metrics}
\label{sec:evaluation_metrics}
To assess the quality of LLM interfaces, we adopt a comprehensive set of evaluation metrics adapted from \citet{nielsen2012usability} and \citet{hartmann2008towards}, capturing three core dimensions of user perception: functional, interactive, and emotional. \textit{Functional Perception} includes \textbf{Query-Interface Consistency (QIC)}, which evaluates how well the generated interface aligns with and fulfills the user's query intent~\citep{duan2025systematic}, and \textbf{Task Efficiency (TaskEff)}, which measures how efficiently users can achieve their goals with minimal effort or time~\citep{nielsen2012usability, duan2025systematic}. \textit{Interactive Perception} comprises \textbf{Usability}, assessing interface clarity and actionable structure~\citep{hartmann2008towards,nielsen2012usability}; \textbf{Learnability}, indicating how easily new users can begin using the interface without prior experience~\citep{nielsen2012usability}; and \textbf{Information Clarity (IC)}, which evaluates information organization, readability, and interpretability~\citep{hartmann2008towards, Cao_2025}. Finally, \textit{Emotional Perception} covers \textbf{Aesthetic or Stylistic Appeal (ASA)}, reflecting the visual consistency and attractiveness of the design~\citep{hartmann2008towards, Duan_2024}, and \textbf{Interaction Experience Satisfaction (IES)}, capturing the user's overall satisfaction and engagement with the interface~\citep{duan2025systematic}.
This enables a comprehensive assessment of user experience by tracing the full perceptual process—``how users understand the interface'' $\rightarrow$ ``how they operate it'' $\rightarrow$ ``how they emotionally respond''.
Instead of using the traditional Likert scale, we adopt a pairwise comparison approach, following \cite{zheng2023lmsyschat1m, si2024design2code}. That is, for each query, we present two interfaces to human annotators and ask for their preferences on all seven dimensions, as well as their overall preferences.

\subsection{Human Evaluation}
\label{sec:human_study}

We conducted a pairwise human evaluation study on Prolific\footnote{\url{https://app.prolific.com}}. Each evaluation instance consisted of a user query paired with two UI outputs (Example 1 and Example 2) generated by different methods. Annotators were asked to judge which output better satisfied seven evaluation dimensions as well as overall quality, selecting among “Example 1 wins,” “Example 2 wins,” or “Tie.” Each instance was evaluated by three annotators, and we aggregated their responses via majority voting to obtain a final decision. Despite the inherent subjectivity of interface evaluation, inter-annotator agreement measured by Fleiss’ Kappa~\citep{landis1977measurement} reached 0.525, indicating a moderate level of consistency among annotators.

Our study involved a total of 428 unique annotators, who were compensated at a rate of \$16/hour. All participants were native English speakers based in the United States and regular users of AI chatbot systems (e.g., ChatGPT). They were experienced annotators, each having completed over 1,000 prior tasks with an approval rate exceeding 90\%. All held at least a bachelor’s degree and were employed either part-time or full-time. The age distribution of annotators was as follows: 18–24 (5.8\%), 25–34 (29.4\%), 35–44 (31.1\%), 45–54 (21.0\%), 55–64 (10.3\%), and 65+ (2.3\%).

%% file: figures/2-method.tex
\begin{figure*}[t]
  \centering
\includegraphics[width=0.99\linewidth]{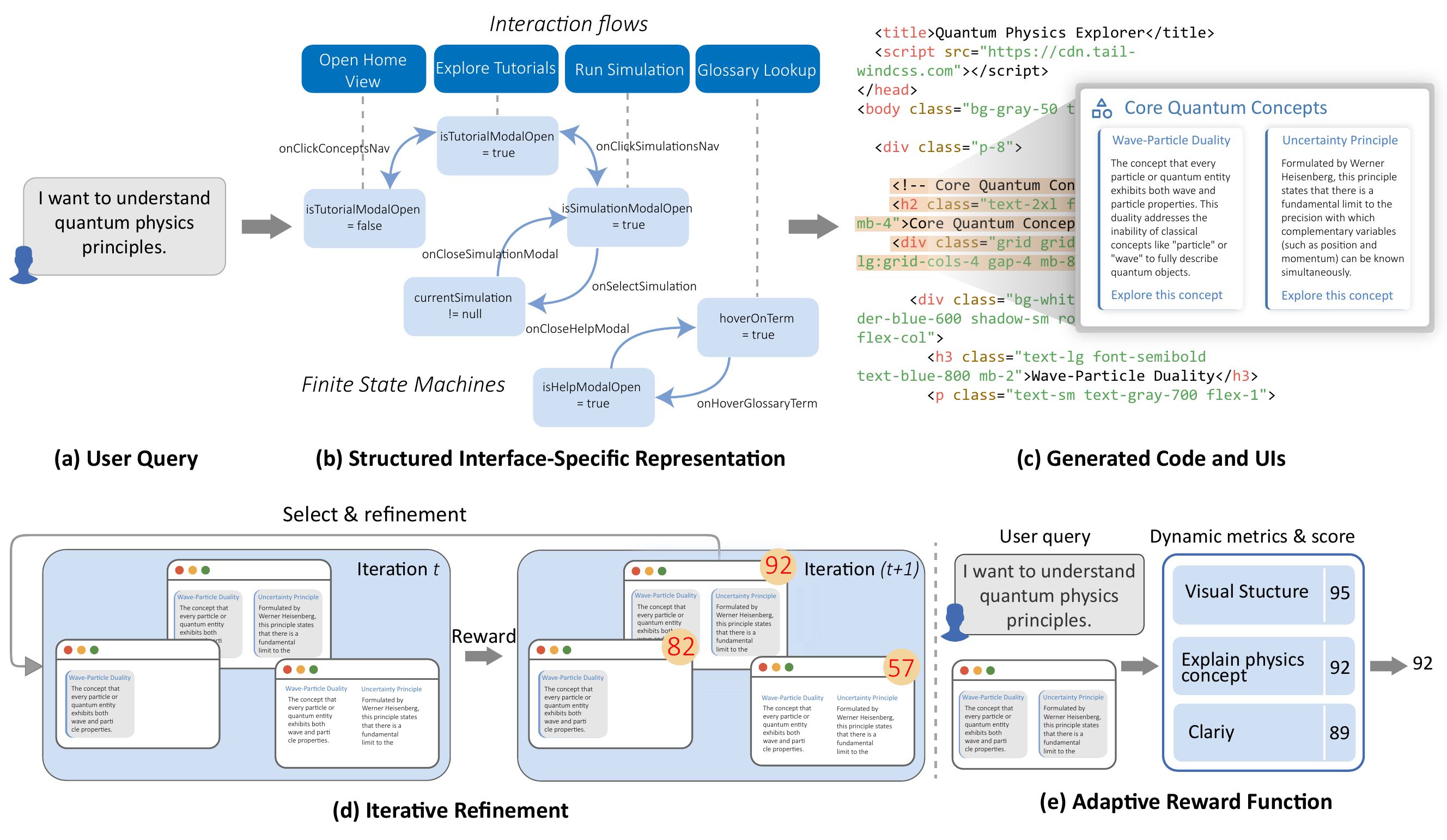}
  \caption{\textbf{Generative Interfaces infrastructure}: (a) User queries are first converted into (b) structured interface-specific representations that model interaction flows and component dependencies. This structured representation guides the generation of (c) functional code and user interfaces. The system employs (d) iterative refinement with (e) adaptive reward functions containing query-specific evaluation rubrics.}
  \label{fig:method}
\end{figure*}

%% file: files/4-exp.tex
\section{Experimental Results}
\input{tables/1-human_evaluation_results}

\noindent
\textbf{Implementation details}
Our system is built on OpenCanvas\footnote{\url{https://github.com/langchain-ai/open-canvas}} and uses Claude 3.7~\citep{anthropicclaude37} as the default backbone LLM, given its strong performance in UI code generation~\citep{si2024design2code, li2024sketch2code}. We refer to our approach as \textbf{GenUI} and compare it against two baselines: 
\textbf{(I) \textit{Conversational UI (ConvUI):}} A traditional chat interface using either GPT-4o~\citep{openai2024gpt4ocard} or Claude 3.7~\citep{anthropicclaude37}. To reduce potential bias in human evaluation, we present a unified chat interface without disclosing the underlying model. For Claude 3.7, 26\% of responses include artifact generation. We remove the artifacts and retain only the textual output to ensure a clean and fair comparison with other conversational systems.  
\textbf{(II) \textit{Instructed UI (IUI):}} An interface generated by Claude 3.7 when explicitly prompted (query + ``\emph{Please help me solve it with UI}''). This prompt consistently triggers artifact generation, and the resulting artifact is taken as the system output.

\subsection{Main Results and Findings}

\noindent
\textbf{Conversational \textit{vs.} Generative Interfaces}
As shown in Table~\ref{tab:main_human}, GenUI consistently outperforms ConvUI across all evaluation dimensions. 
Interestingly, ConvUI (GPT-4o) performs more competitively than ConvUI (Claude 3.7), suggesting that well-structured textual responses can still be effective in specific scenarios.
Compared to ConvUI (Claude 3.7), GenUI achieves the most significant gains in ASA (+86.0\%) and IES (+81.0\%). Overall, its emotional appeal and interactive functionality are the primary drivers of its superior performance, resulting in an 84.0\% win rate over ConvUI (Claude 3.7). These findings suggest that users clearly prefer GenUI for most queries.

User comments further support this finding. For example, one noted:
``\textit{GenUI provides the requested information in an \textbf{easy-to-understand manner}, laying out everything requested and anticipating what else may be needed.}''
A small number of users did express a preference for the familiarity of traditional ConvUIs, as one remarked (see interface examples in Appendix Figure~\ref{fig:example_stratege}):
``\textit{Chatbot interface is \textbf{most people know already}, while GenUI is a somewhat complex and \textbf{unfamiliar} app.}''
This suggests that some users remain attached to familiar formats due to habits or ease.
However, such a preference did not override the broader recognition of GenUI’s objective advantages, indicating strong potential for user adaptation and adoption in real-world deployments.

\input{figures/3_win_rate_comparison}

\noindent
\textbf{Domain Analysis}
As shown in Table~\ref{fig:exp-scenarios-human-study}, preferences for GenUI vary by domain. Users strongly favored GenUI in \textit{Data Analysis \& Visualization} (93.8\%) and \textit{Business Strategy \& Operations} (87.5\%), where tasks typically involve interpreting large amounts of structured information. By contrast, in \textit{Advanced AI/ML Applications}, GenUI received 50.0\% of preferences, suggesting that traditional linear text explanations remain effective in math-heavy contexts. Overall, these results indicate that domains characterized by complex information benefit most from GenUIs, whereas ConvUIs are still suitable for domains that rely on straightforward explanations.

\input{figures/9_human_preference_breakdown}

\noindent
\textbf{Query Analysis}
As shown in Figure~\ref{fig:task-prompt-preference-breakdown}, GenUI receives stronger preferences for certain query characteristics. It is particularly favored in interactive tasks (80.0\%), underscoring the advantages of generative interfaces in scenarios where interaction is essential for task completion.
In general conversations, users also show a clear preference for GenUI over ConvUI (73.0\% \textit{vs.} 23.0\%).
When comparing query detail level, GenUI is preferred more for detailed queries (80.0\%) than for concise ones (73.0\%), likely because simple conversational responses sometimes sufficiently address short queries, whereas GenUI may introduce unnecessary complexity.

\input{tables/6-ablation_human}
\subsection{Ablation Study}

\textbf{(I) Our Pipeline \textit{vs.} Direct Instruct:}
We compare our framework against IUI: directly instructing Claude 3.7 to generate a web interface with the artifact feature enabled, representing a highly engineered baseline.
Our system outperforms this strong baseline, achieving a 58.0\% higher win rate (Table~\ref{tab:main_human}).
Among the baselines, IUI shows better performance in emotional perception dimensions such as ASA, but it still lags behind GenUI overall.
\textbf{(II) Natural Language \textit{vs.} Structured Representation:}
The natural language version provides a descriptive explanation of the UI based on the user query, without employing structured representations to define interface states formally.
As shown in Table~\ref{tab:ablation_human} (Row $1$ \textit{vs.} Row $2$), structured representations outperform natural language, improving the win rate from 13\% to 17\% overall.
\textbf{(III) One-shot Generation \textit{vs.} Iterative Refinement:}
As shown in Table~\ref{tab:ablation_human} (Row $2$ \textit{vs.} Row $3$), the iterative refinement process yields consistent improvements on human preference across all perception dimensions, resulting in a notable +14.0\% overall win rate improvement compared to one-shot generation. Figure~\ref{fig:performance-iteration} further illustrates that each refinement round leads to a clear performance boost, with average LLM-based reward scores increasing by +1.2\% and +4.9\%, respectively. 
We illustrate an example of such iterative improvement in Appendix Figure~\ref{fig:example_ci_iteration}, where each iteration incrementally enhances layout efficiency, usability, and user guidance, ultimately leading to a more informative and user-friendly interface through structured refinement.
\textbf{(IV) Static \textit{vs.} Adaptive Reward Function:}
Table~\ref{tab:ablation_human} (Row $3$) highlights the effect of dynamic reward functions, which differ from the full version only by replacing adaptive scoring with a static baseline. 
The absence of dynamic rewards results in a 17.0\% drop in overall win rate, with performance declining across all seven evaluation metrics. 
This comparison highlights the importance of dynamically adjusting evaluation criteria to capture the task-specific requirements inherent in each query instead of generic, fixed heuristics.

\subsection{Human Preference Analysis}
\label{sec:human_comments_analysis}
\input{figures/11_human_comment}

To better understand the factors underlying human annotator preferences, we collected fine-grained textual justifications for each perception dimension in 40\% of the pairwise comparisons, and overall comments for the remaining 60\%. Following the methodology of~\citet{lam2024concept}, we used Claude 3.7 to systematically extract high-level semantic concepts from these qualitative responses.
The resulting comments were then clustered into semantic themes identified by the LLM (Table~\ref{table:exp-human-comment-distribution}).
This analysis allows us to pinpoint the key factors shaping user preferences beyond surface-level considerations such as visual aesthetics and engagement.
Finally, we computed preference distributions between generative and conversational interfaces within each semantic themes.

\paragraph{Cognitive Offloading}~\citep{risko2016cognitive} emerges through user comments as a subtler yet deeper driver.
78.5\% of users mentioning \textit{Cognitive Load \& Intuition} preferred GenUI.
For instance, in designing a continuing education program for healthcare professionals, a user noted:
“\textit{This type of information analysis is very complex \dots GenUI helps to \textbf{break down the categories into manageable steps} \dots makes the complex information easier to process.}”
This illustrates how GenUI’s interface acts as an external cognitive aid to break down information.
However, in easier scenarios such as designing a high-school mathematics curriculum, ConvUI was preferred because it “\textit{clearly and informatively illustrates the steps}”.
In summary, GenUI excels in \textbf{\textit{complex, concept-heavy scenarios}} where cognitive offloading facilitates understanding. In contrast, ConvUI outperforms for easy and basic \textbf{\textit{“how-to” queries}} where additional tools impose unnecessary cognitive load.

\paragraph{Perceived Usability and Trust}
Among the user comments related to the “\textit{Perceived Credibility \& Professionalism}” dimension, 86.5\% preferred the GenUI. Users consistently described GenUI as more authoritative, credible, and professional.
For example, in response to the query “\textit{How do I conduct market research?}”, users commented:
    “\textit{GenUI is more professionally written}”, 
    “\textit{It offers out the more sound advice}”, and
    “\textit{It is the better discernment.}”
Notably, this perception of professionalism does not stem solely from the content itself. In fact, many users acknowledged that both interfaces provided reasonable answers to the query (\textit{e.g.}, “\textit{Both answer the prompt reasonably well}”). What sets GenUI apart is its presentation: through modular layouts, clear hierarchies, visual anchors, and polished formatting, it delivers the information in a more organized manner.

\subsection{Real-User Query Evaluation}
To further test generalizability, we conducted an additional study using real-user queries, where participants first provided five of their typical LLM queries \citep{bassignana-etal-2025-ai} and then compared ConvUI and GenUI on these self-authored tasks (see Appendix \ref{appendix: Real-User Query Evaluation} for more details). The collected queries encompass a diverse range of organically occurring user needs, from informational and planning tasks (e.g., travel, shopping, education) to personal, emotional, and creative requests. More importantly, this setup better captures authentic usage contexts and reduces query–annotator mismatch. Across 380 queries from 76 participants, GenUI demonstrated a clear advantage (50.8\% win, 8.2\% tie, 41.1\% loss) compared to ConvUI.
Among all participants, 30.3\% of them strongly preferred GenUI (in $\geq$80\% of cases) even though this is their first time experiencing it, whereas only 18.4\% strongly preferred ConvUI.
The remaining users exhibited more query-dependent choices.
Notably, for underrepresented domains in the UIX benchmark, such as personal and emotional tasks, GenUI still stands out, as users praised ``\emph{its visual appeal, interactivity, and personalized, tool-like experience that felt more engaging and immersive}''. Overall, this study reinforces our main findings.

%% file: tables/1-human_evaluation_results.tex
\begin{table*}[t]
\centering
\small
\renewcommand\arraystretch{1.2}
\setlength{\tabcolsep}{2.3pt}
\begin{tabular}{l|c|cc|ccc|cc|c}
\hline

\hline

\hline
 
\hline
\multirow{2}{*}{\textbf{Framework}} & \multirow{2}{*}{\textbf{Status}}
& \multicolumn{2}{c|}{\textit{\textbf{Functional}}} 
& \multicolumn{3}{c|}{\textit{\textbf{Interactive}}} 
& \multicolumn{2}{c|}{\textit{\textbf{Emotional}}} 
& \multirow{2}{*}{\textbf{Overall}} \\
 & & QIC &TaskEff 
& Usability & Learnability & IC 
& ASA & IES 
& \\
\hline
                \multirow{3}{*}{ConvUI (Claude 3.7) \textit{vs.} GenUI} & ConvUI & 11\% & 14\% & 13\% & 10\% & 9\% & 3\% & 6\% & 12\% \\
                & Tie & 6\% & 5\% & 4\% & 6\% & 6\% & 8\% & 7\% & 4\% \\
                \rowcolor[gray]{.9}
                \cellcolor{white} &  GenUI & 83\% & 81\% & 83\% & 84\% & 85\% & 89\% & 87\% & 84\% \\
\hline
                \multirow{3}{*}{ConvUI (GPT-4o) \textit{vs.} GenUI} & ConvUI & 32\% & 41\% & 28\% & 35\% & 38\% & 13\% & 24\% & 30\%\\
                & Tie & 11\% & 5\% & 7\% & 10\% & 8\% & 7\% & 6\% & 1\%\\
                \rowcolor[gray]{.9}
                \cellcolor{white} & GenUI & 57\% & 54\% & 65\% & 55\% & 54\% & 80\% & 70\% & 69\% \\
\hline
                \multirow{3}{*}{IUI \textit{vs.} GenUI} & IUI & 13\% & 17\% & 16\% & 14\% & 16\% & 20\% & 14\% & 17\% \\
                & Tie & 18\% & 13\% & 18\% & 20\% & 15\% & 5\% & 15\% & 8\% \\
                \rowcolor[gray]{.9}
                \cellcolor{white} & GenUI & 69\% & 70\% & 66\% & 66\% & 69\% & 75\% & 71\% & 75\% \\

\hline

\hline

\hline

\hline
\end{tabular}
\caption{\textbf{Human Evaluation of UI Framework.} 
Win, tie, and loss percentages of UI variants compared to our system (GenUI) across different perception dimensions: functional, interactive, and emotional.
}
\label{tab:main_human}
\end{table*}

%% file: figures/3_win_rate_comparison.tex

\begin{table}[t]
\centering
\setlength{\tabcolsep}{3pt}
\renewcommand{\arraystretch}{1.05}
\resizebox{\linewidth}{!}{
\begin{tabular}{l|c}
\hline

\hline

\hline

\hline
\textbf{Domain} & \textbf{GenUI(\%)} \\
\hline
Data Analysis \& Visualization & 93.8 \\
Language Translation & 87.5 \\
Business Strategy \& Operations & 87.5 \\
Education \& Career Development & 83.3 \\
Academic Research \& Writing & 79.2 \\
Content Creation \& Communication & 75.0 \\
Digital Marketing \& SEO & 75.0 \\
DevOps \& Cloud Infrastructure & 75.0 \\
Web \& Mobile App Development & 70.8 \\
Advanced AI/ML Applications & 50.0 \\
\hline

\hline

\hline

\hline
\end{tabular}}
\caption{The ratio of GenUI being preferred across 10 query topics~\citep{tamkin2024clio}.}
\label{fig:exp-scenarios-human-study}
\end{table}

%% file: figures/9_human_preference_breakdown.tex
\begin{figure*}[t]
  \centering
  \begin{subfigure}[b]{0.45\textwidth}
    \centering
    \includegraphics[width=\linewidth]{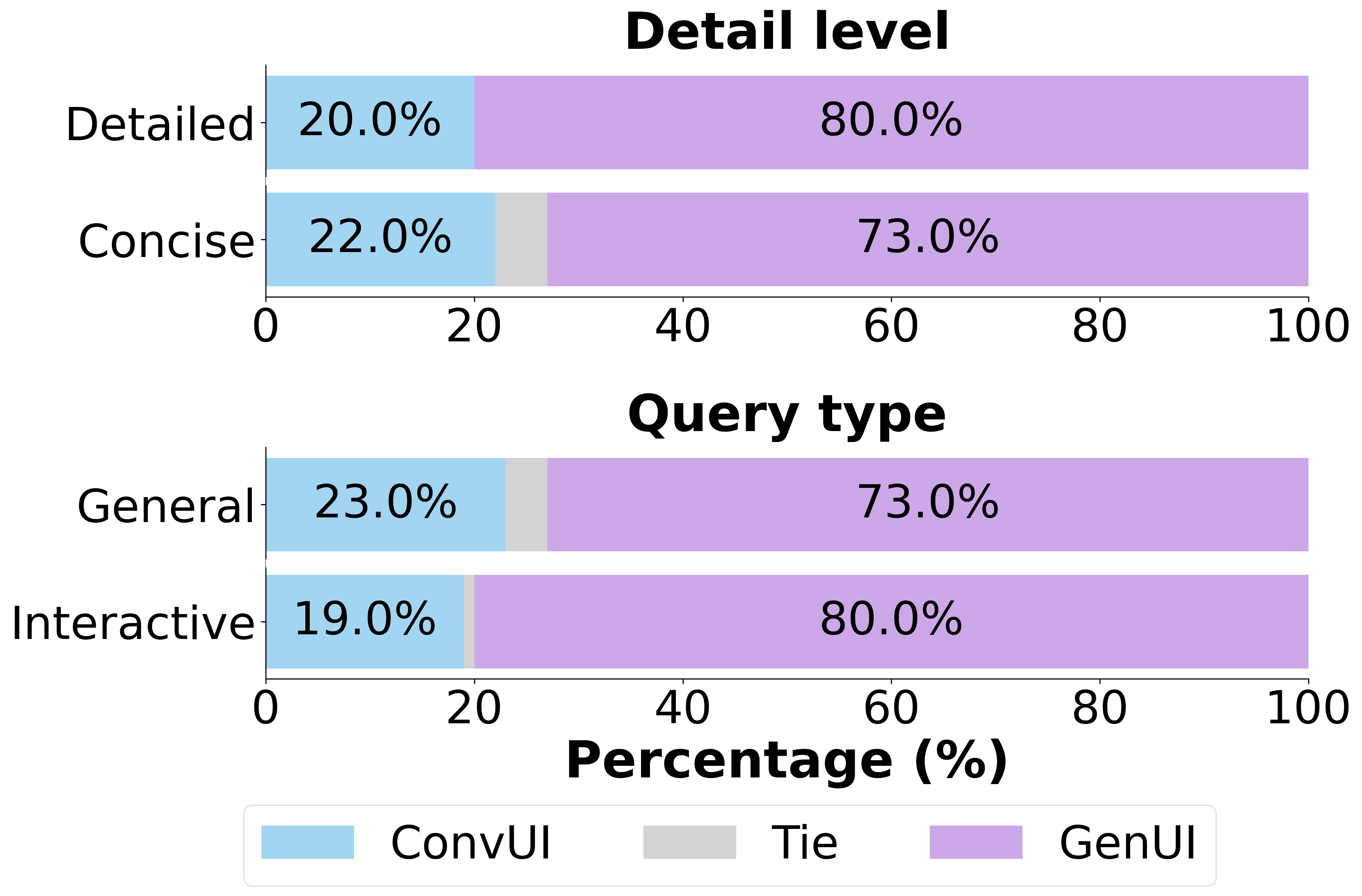}
    \caption{\textbf{Breakdown of query detail level and type.}}
    \label{fig:task-prompt-preference-breakdown}
  \end{subfigure}
  \hfill
  \begin{subfigure}[b]{0.45\textwidth}
    \centering
    \includegraphics[width=\linewidth]{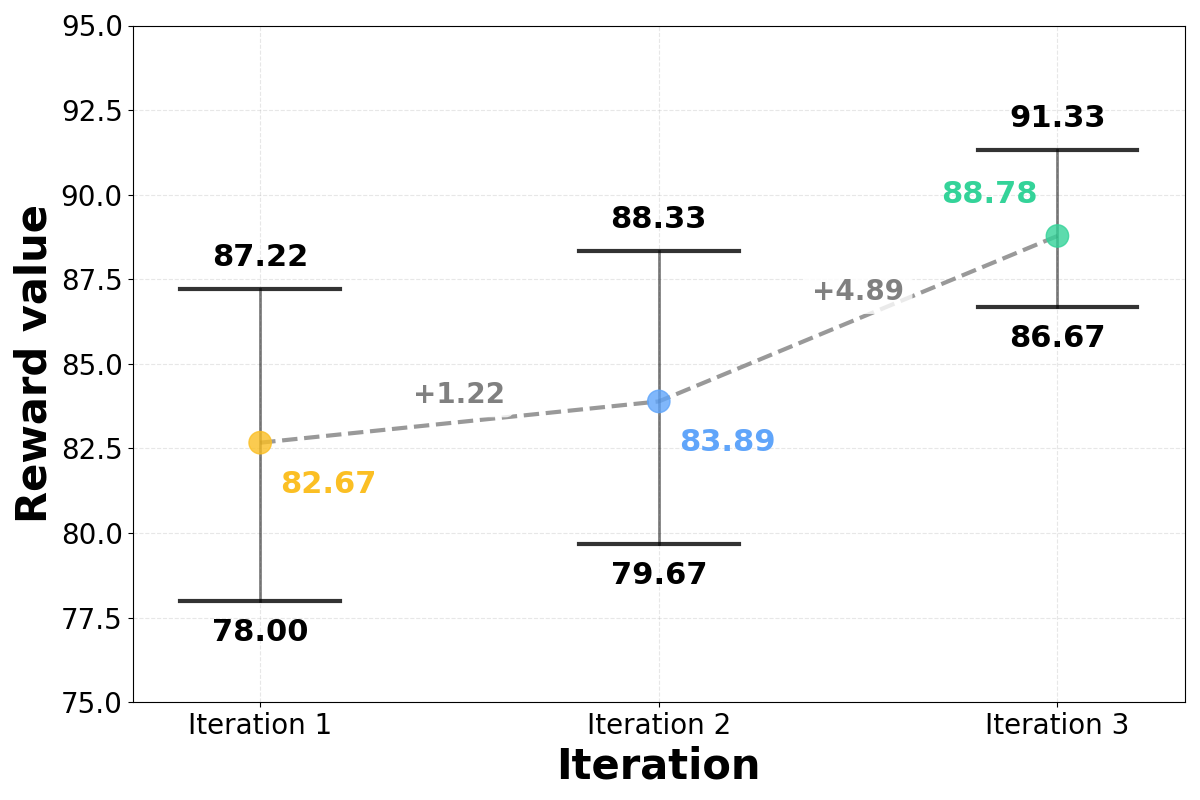}
    \caption{\textbf{Ablation on number of iterations.}
    }
    \label{fig:performance-iteration}
  \end{subfigure}
  \caption{Human evaluation results comparing GenUIs and ConvUIs. (a) User preference breakdown by query type and detail level. (b) Performance improvement across iterative interactions.}
  \label{fig:human-eval-combined}
\end{figure*}

%% file: tables/6-ablation_human.tex
\begin{table*}[t]
\centering
\small
\renewcommand\arraystretch{1.2}
\setlength{\tabcolsep}{2pt}
\begin{tabular}{ccc|c|cc|ccc|cc|c}
\hline

\hline

\hline
 
\hline
\textbf{Reward} & \textbf{Generation} & \textbf{Repre-} & \multirow{2}{*}{\textbf{Status}}
& \multicolumn{2}{c|}{\textit{\textbf{Functional}}} 
& \multicolumn{3}{c|}{\textit{\textbf{Interactive}}} 
& \multicolumn{2}{c|}{\textit{\textbf{Emotional}}} 
& \multirow{2}{*}{\textbf{Overall}} \\
\textbf{design} & \textbf{paradigm} & \textbf{sentation} &
& QIC & TaskEff 
& Usability & Learnability & IC 
& ASA & IES 
& \\
\hline
\multicolumn{12}{l}{\textit{Full GenUI: Adaptive, Iterative, Structured}} \\ \hline
\multirow{3}{*}{Static} &   \multirow{3}{*}{One-shot} & \multirow{3}{*}{Natural}
& Win & 8\%  & 16\% & 11\% & 19\% & 15\% & 10\% & 13\% & 13\% \\
&&& Tie  & 20\% & 20\% & 22\% & 16\% & 15\% & 13\% & 15\% & 5\% \\
 \rowcolor[gray]{.9} \cellcolor{white}&\cellcolor{white}&\cellcolor{white}& Loss & 72\% & 64\% & 67\% & 65\% & 70\% & 77\% & 72\% & 82\% \\
\hline

\multirow{3}{*}{Static}  & \multirow{3}{*}{One-shot} & \multirow{3}{*}{Structured}  
& Win & 11\% & 18\% & 18\% & 18\% & 16\% & 15\% & 14\% & 17\% \\
&&& Tie  & 20\% & 12\% & 12\% & 15\% & 10\% & 10\% & 13\% & 5\% \\
 \rowcolor[gray]{.9} \cellcolor{white}&\cellcolor{white}&\cellcolor{white}& Loss & 69\% & 70\% & 70\% & 67\% & 74\% & 75\% & 73\% & 78\% \\
\hline

\multirow{3}{*}{Static} & \multirow{3}{*}{Iterative} & \multirow{3}{*}{Structured}  
& Win & 28\% & 30\% & 30\% & 27\% & 27\% & 34\% & 27\% & 31\% \\
&&& Tie  & 32\% & 26\% & 24\% & 30\% & 27\% & 17\% & 28\% & 15\% \\
 \rowcolor[gray]{.9} \cellcolor{white}&\cellcolor{white}&\cellcolor{white}& Loss & 40\% & 44\% & 46\% & 43\% & 46\% & 49\% & 45\% & 54\% \\
\hline

\hline

\hline

\hline
\end{tabular}
\caption{\textbf{Ablation study.} 
The control group is the full GenUI framework (\textit{adaptive} reward, \textit{iterative} generation, and \textit{structured} representation). 
All ablations are compared against this full version, where “Loss” indicates that GenUI outperforms the variant.
Note that ``\textit{Static}'' refers to static reward design, ``\textit{One-shot}" denotes generation without refinement, and ``\textit{Natural}" indicates natural language representations.
}
\label{tab:ablation_human}
\end{table*}

%% file: figures/11_human_comment.tex

\begin{table}[t]
\centering
\setlength{\tabcolsep}{3pt}
\renewcommand{\arraystretch}{1.05}
\resizebox{\linewidth}{!}{
\begin{tabular}{l|c}
\hline

\hline

\hline

\hline
\textbf{Concept} & \makecell{\textbf{GenUI}\\\textbf{(\%)}}  \\
\hline
Visual Aesthetics \& Engagement (23.4\%) & 83.3 \\
Information Organization \& Accessibility (14.9\%) & 87.4 \\
Cognitive Load \& Intuition (14.5\%) & 78.5 \\
Actionability \& Practical Utility (10.7\%) & 81.8 \\
Information Richness \& Comprehensiveness (10.4\%) & 82.5 \\
Guidance \& Learning Support (9.7\%)  & 74.5 \\
Content Relevance \& Efficiency (6.8\%)  & 59.0 \\
Interactive Experience Quality (5.5\%)  & 89.7 \\
Perceived Credibility \& Professionalism (3.3\%)  & 86.5 \\
Others (1.0\%)  & 71.9 \\
\hline

\hline

\hline

\hline
\end{tabular}}
\caption{\textbf{Concept distribution.} We show the distribution of high-level concepts extracted from user comments using the pipeline described in Sec.~\ref{sec:human_comments_analysis}. For each concept, we show the ratio of GenUIs being preferred.}
\label{table:exp-human-comment-distribution}
\end{table}

%% file: files/2-related.tex
\section{Related work}

\noindent
\textbf{Context-Aware and Adaptive Interface}
Context-aware interfaces have been widely explored since the rise of ubiquitous computing, aiming to improve usability, reduce cognitive load, and better support user goals~\citep{dey2000towards, horvitz1999principles, theng2008ubiquitous}. As computing systems have become more complex and pervasive, the ability to adjust interfaces dynamically has been critical for creating more effective and accessible user experiences~\citep{gajos2004supple, gajos2007automatically, nichols2002generating, nichols2006uniform, nichols2006huddle}. Prior systems often adapted functionality through a finite set of predefined states. While effective in constrained settings, these approaches faced challenges with scalability and sometimes reduced predictability and user control~\citep{findlater2009design}.
Recent advances in LLMs have enabled new forms of adaptive interfaces that dynamically generate interface elements in response to user prompts~\citep{wu2022promptchainer, dibia2023lida, cha2024shared, cheng2024biscuit, nandy2024bespoke}, marking a shift from static outputs toward model-driven, interactive systems.

To improve interaction efficiency between humans and LLMs, prior studies ~\citep{jiang2023graphologue,ma2024beyond,ross2023programmer} have proposed combining text-based ConvUIs with Graphical User Interfaces (GUIs). For example, OpenAI Canvas enables users to directly edit documents and code on a canvas, avoiding repeated prompt inputs; Graphologue~\citep{jiang2023graphologue} transforms lengthy and complex LLM responses into graphical diagrams to support information exploration and question answering.
However, although these approaches leverage LLMs to generate displayed content, the UIs they employ are predesigned. 
In contrast, GenerativeGUI~\citep{hojo2025generativegui} explores the usability of dynamically generated interfaces in clarifying question (CQ) interactions. ClarifyGPT~\citep{mu2023clarifygpt} also introduces CQs, but in the narrower domain of code generation.
Beyond CQ scenarios, DynaVis~\citep{vaithilingam2024dynavis} proposes a system that combines natural language with dynamically synthesized UI widgets to support chart editing tasks, without exploring broader, general-purpose scenarios.
Unlike prior systems that modify fixed UI components~\citep{Cao_2025}, our framework generates complete interfaces customized to diverse user queries.

\noindent
\textbf{Automatic UI Generation}
This direction has evolved from early vision-based approaches to reverse engineering mobile interfaces \citep{nguyen2015reverse} to neural network-based end-to-end synthesis systems \citep{beltramelli2018pix2code, Robinson2019Sketch2codeGA, 8741736}. Recent advances in LLMs have substantially improved UI generation from natural language descriptions \citep{laurenon2024unlocking}, screenshots \citep{si2024design2code}, and sketches \citep{li2024sketch2code} and iterative refinement via LLM-generated feedback \citep{li2024sketch2code}.
Alternatively, our work requires no UI specifications from users.
More fine-grained control of the UI code generation process encompasses diverse intermediate representation approaches: (I) graph-based representation to capture hierarchical relationships and dependencies between UI elements \citep{jiang2024graph4gui}, (II) UI grammar \citep{kong2008adaptive} to help LLMs for more intuitive and precise layout description \citep{lu2023uilayout}, and (III) data schema-driven UI specification synthesis to guide subsequent generation processes \citep{Cao_2025}.
Similarly, our framework employs interaction flows and finite state machines to model the reaction to user actions and the evolution of interfaces.

%% file: files/5-con.tex
\section{Conclusion}
We introduce \textbf{Generative Interfaces} for Language Models, a paradigm in which LLMs proactively generate adaptive, interactive interfaces to better support complex user goals.  
Our evaluation demonstrates clear advantages over traditional conversational approaches, particularly in structured and information-dense tasks.  
Our findings further clarify when generative interfaces are most effective and when conversational formats remain competitive.
Future directions include integrating multimodal input, domain-specific templates, and collaborative multi-user environments.

\section*{Limitations}  
First, the system only supports HTML/JavaScript frontends without backend logic, which restricts the complexity of generated interfaces. As tasks grow more complex, more expressive representations beyond interaction flows and finite state machines may be needed.  
Second, the iterative refinement process introduces latency of up to several minutes, which may be undesirable in real-time settings. Advances in model efficiency and infrastructure could help mitigate this issue.  
Third, the system generates interfaces for all queries, even when interaction is unnecessary. Future work could incorporate a classifier to determine whether an input requires interaction in context and selectively invoke the generative UI system.

\section*{Ethical Considerations}
Generative interfaces may create accessibility barriers for users relying on assistive technologies.
By shaping user interpretation and decision-making, they may raise risks of unintended persuasion or biased framing, particularly in high-stakes contexts.
The polished, tool-like presentation of outputs may also increase user trust and lead to overconfidence, while the outputs might contain misinformation and hallucinations.
Addressing these issues will require improved transparency and oversight mechanisms to ensure that generative interfaces remain both usable and trustworthy.

\section*{Acknowledgments}
We thank anonymous reviewers and SALT Lab members for their valuable feedback on this work.
This work is supported by ONR N000142412532, NSF IIS 2247357, and Stanford HAI.

%% file: files/appendix.tex
\section{Prompt Suite}
\label{appendix:prompt_suite}
To evaluate system performance across realistic user intents, we curated a prompt suite covering ten practical domains:
\begin{itemize}
    \item \texttt{Web \& Mobile App Development}, 
    \item \texttt{Content Creation \& Communication}, 
    \item \texttt{Academic Research \& Writing}, 
    \item \texttt{Education \& Career Development}, 
    \item \texttt{Advanced AI/ML Applications}, 
    \item \texttt{Business Strategy \& Operations}, 
    \item \texttt{Language Translation}, 
    \item \texttt{DevOps \& Cloud Infrastructure}, 
    \item \texttt{Digital Marketing \& SEO}, and 
    \item \texttt{Data Analysis \& Visualization}.
\end{itemize}

Each prompt belongs to one of four quadrants based on \emph{detail level} (concise \textit{vs.}\ detailed) and \emph{type} (general \textit{vs.}\ interactive), ensuring coverage of diverse user tasks and complexity levels. Here are some examples:

\begin{itemize}
    \item \textbf{Concise \& General:} \textit{``How can I learn piano effectively?''}

    \item \textbf{Concise \& Interactive:} \textit{``I want to create an infographic about water conservation.''}
    \item \textbf{Detailed \& General:} \textit{``I'm writing a dissertation on the psychological effects of social media use among teenagers. I've collected survey and interview data but am struggling to integrate them in the analysis chapter. What methodological approach should I use to synthesize these data types rigorously?''}

    \item \textbf{Detailed \& Interactive:} \textit{``I'm developing a website for a local bookstore where customers can browse inventory, register for book club meetings, and sign up for our newsletter. I want a cozy design but have no coding experience. The inventory is in Excel and updates weekly. What’s the best approach to build this site?''}
\end{itemize}

\section{LLM Evaluation}
\label{sec:llm_evaluation}

\input{tables/2-llm_evaluation_results}

User-centered evaluation remains the gold standard for UI assessment due to interfaces' fundamental purpose of facilitating human interaction and operation \citep{hartmann2008towards, duan2025systematic, Cao_2025}. However, generative interfaces requiring real-time synthesis and rapid iterative refinement cannot depend on user feedback, necessitating robust automatic evaluation frameworks.

Early approaches employed manually crafted behavioral prediction metrics \citep{Lee_2020}, though these methods demonstrated limited generalizability and required substantial domain expertise. Recent research has increasingly leveraged LLMs for UI assessment, with \citet{Duan_2024} employing LLMs to generate design feedback and quality ratings with bounding box annotations. \citet{jeon2025gfocus} extends this paradigm to persuasiveness evaluation, achieving meaningful correlation with empirical A/B testing outcomes.
In this work, we ask LLMs to judge the same dimensions that we ask human annotators, including some previously human-exclusive metrics such as task efficiency and learnability.

Specifically, we use a listwise ranker \citep{liu2023dynamic} to evaluate different interface variants of the same user query by presenting the LLM (Claude 3.7) with UI codes and screenshots, where the LLM assigns scores ranging from $0$ to $100$ for each evaluation dimension.
We compare LLM evaluation scores with pairwise annotations from humans, which yields an agreement rate of $69.0\%$. While it suggests LLM as a reliable proxy for convenient and scalable evaluations, we also observe common issues like length bias \citep{dubois2024lengthcontrolled} which might favor GenUI.

\begin{table*}[h]
\centering
\renewcommand{\arraystretch}{1.3}
\begin{tabular}{p{4.2cm} p{9cm}}
\hline

\hline

\hline

\hline
\textbf{Term / Symbol} & \textbf{Definition} \\
\midrule
\textbf{Interaction Flow} & A high-level abstraction over user interaction sequences, modeling task progression as transitions across interface views. \\
\midrule
$\mathcal{G} = (\mathcal{V}, \mathcal{T})$ & Directed graph structure of the interaction flow: $\mathcal{V}$ is the set of views or subgoals, and $\mathcal{T}$ is the set of transitions between them. \\

$\mathcal{V}$ & Nodes in the interaction graph, each representing a specific UI view or a subgoal in the task. \\

$\mathcal{T}$ & Directed edges indicating possible user-triggered transitions between views, such as button clicks or link navigation. \\
\midrule
\textbf{Finite State Machine (FSM)} & A formalism used to describe the behavior and state transitions of individual UI components based on user interactions. \\
\midrule
$\mathcal{M} = (\mathcal{S}, \mathcal{E}, \delta, s_0)$ & Formal definition of an FSM: $\mathcal{S}$ is the state set, $\mathcal{E}$ the event set, $\delta$ the transition function, and $s_0$ the initial state. \\

$\mathcal{S}$ & Set of all possible atomic interface states (\textit{e.g.}, \texttt{isModalOpen=true}, \texttt{activeTab=2}). \\

$\mathcal{E}$ & Set of discrete user-triggered events such as \texttt{click}, \texttt{hover}, \texttt{input}, etc. \\

$\delta$ & Transition function $\delta: \mathcal{S} \times \mathcal{E} \rightarrow \mathcal{S}$, defining how a component’s state evolves given an event. \\

$s_0$ & The initial state of the UI component when the interface is first rendered. \\

\hline

\hline

\hline

\hline
\end{tabular}
\caption{Glossary of concepts and formal symbols used in structured interface-specific representation.}
\label{tab:glossary}
\end{table*}

\section{Representation: Natural language \textit{vs.} Structured}
\label{appendix:representation}

To present a more fine-grained comparison, we showcase two distinct representations of the same user intent. The user prompt used here is: \begin{quote}
    ``\textit{I want to understand quantum physics principles.}''
\end{quote}
A natural language representation includes the goal, salient features, technical requirements, and user preferences, which are expressed through multiple descriptive fields. This format provides rich detail about the UI requirements without imposing any constraints on the interface states or their transitions.

\begin{tcolorbox}[title={\textbf{\small  Natural language representation}}, colback=whitesmoke, colframe=royalblue(web), boxrule=2pt, arc=0mm, breakable]
{\scriptsize
\begin{lstlisting}[style=mystyle]
{
  "mainGoal": "Create an interactive learning interface for understanding quantum physics principles.",
  "keyFeatures": [
    "Step-by-step tutorials on key quantum physics concepts",
    "Interactive simulations demonstrating quantum mechanics principles",
    "Visual aids such as diagrams and animations to enhance comprehension",
    "Quizzes and assessments to test understanding and reinforce learning",
    "Discussion forums for peer interaction and support"
  ],
  "technicalRequirements": [
    ...
  ],
...
}
\end{lstlisting}
}
\end{tcolorbox}
Structured interface-specific representation is state-oriented and descriptive. In table \ref{tab:glossary}, we summarize concepts and symbols used in structured interface-specific representations.
\begin{tcolorbox}[title={\textbf{\small  Structured representation}}, colback=whitesmoke, colframe=royalblue(web), boxrule=2pt, arc=0mm, breakable]
{\scriptsize
\begin{lstlisting}[style=mystyle]
{
  "description": "An interactive educational platform for learning quantum physics principles through tutorials, simulations, quizzes, progress tracking, and discussion forums. The platform offers step-by-step learning paths, visual demonstrations of quantum phenomena, and assessment tools to help users understand complex quantum physics concepts.",
  "metadata": {
    "title": "Quantum Physics Explorer - Interactive Learning Platform",
    "metaDescription": "Learn quantum physics through interactive tutorials, simulations, quizzes, and discussion forums. A comprehensive educational platform for understanding quantum mechanics principles."
  },
  "states": [
    {
      "name": "isMobileMenuOpen",
      "initialValue": "false",
      "description": "Controls the visibility of the mobile navigation menu on smaller screens."
    },
    ...
   "elements": [
    ...
    {
      "id": "helpButton",
      "parentId": "userControls",
      "elementType": "button",
      "content": "Help",
      "className": [
        "text-blue-600",
        "hover:text-blue-800",
        "focus:outline-none",
        "focus:ring-2",
        "focus:ring-blue-500",
        "rounded-full",
        "p-2"
      ],
      "functionality": "Provides access to help resources and tutorials.",
      "attributes": {
        "ariaLabel": "Get help"
      },
      "events": [
        {
          "type": "onClick",
          "handlerDescription": "Opens the help modal with tutorials and resources.",
          "affects": [
            {
              "target": "isHelpModalOpen",
              "action": "updateState",
              "details": "true"
            }
          ]
        }
      ],
      "interactions": {
        "hover": {
          "className": [
            "text-blue-800",
            "bg-blue-50"
          ]
        },
        "focus": {
          "className": [
            "ring-2",
            "ring-blue-500"
          ]
        }
      }
    },
      "flows": [
    {
      "name": "Explore Tutorials",
      "description": "User navigates to and interacts with the tutorials section to learn about quantum physics concepts.",
      "steps": [
        "User scrolls down to the 'Quantum Physics Tutorials' section or clicks on the 'Tutorials' navigation item.",
        ...
      ]
    },
    ...
\end{lstlisting}
}
\end{tcolorbox}

\section{Adaptive Reward Function}
\label{appendix:reward_function}

The reward function consists of multiple evaluation metrics, each defined with four key fields: 
\begin{itemize}
    \item \textbf{\texttt{name}}: The high-level evaluation dimension.
    \item \textbf{\texttt{description}}: A brief explanation of the dimension's purpose.
    \item \textbf{\texttt{criteria}}: A list of granular human-interpretable evaluation checks.
    \item \textbf{\texttt{weight}}: The relative importance of the metric in the aggregated reward.
\end{itemize}
These metrics collectively guide the assessment of the interface from both functional and user-centered perspectives.

\input{figures/examples/5-reward}
For example, for the user query ``\textit{I want to understand quantum physics principles,}” the adaptive reward metric includes a specific criterion stating that ``\textit{Interactive models effectively demonstrate phenomena like wave-particle duality.}”, which provides intent-aware reward signals that move beyond generic usability (Figure~\ref{fig:dynamic_reward}).
In contrast, the static reward approach yields suboptimal results where particle distributions appear as incoherent clusters without proper interference visualization (Figure~\ref{fig:static_reward}).

\begin{tcolorbox}[title={\textbf{\small  Adaptive reward function}}, colback=whitesmoke, colframe=royalblue(web), boxrule=2pt, arc=0mm, breakable]
{\scriptsize
\begin{lstlisting}[style=mystyle]
{
  "name": "generate_metrics",
  "args": {
    "metrics": [
      {
        "name": "Interactive Elements Quality",
        "description": "Measures the quality of user interaction with simulations, quizzes, and other dynamic components.",
        "weight": 0.15,
        "criteria": [
          "Animations and transitions are smooth and non-distracting.",
          "User actions (e.g., answering quiz questions, changing simulation variables) receive clear and immediate feedback.",
          "Interactive components (simulations, quiz buttons) are responsive to user input.",
          "User flows through tutorials, simulations, and quizzes are intuitive.",
          "State changes (e.g., quiz progress, simulation results) are accurately reflected.",
          "Error prevention mechanisms in quizzes (e.g., guiding towards correct answers) are effective."
        ]
      },
    ...
    ]
  }
}
\end{lstlisting}
}
\end{tcolorbox}

On the other hand, static rewards are defined by general UI rubrics without query-specific metrics.
\begin{tcolorbox}[title={\textbf{\small  Static rewawrd}}, colback=whitesmoke, colframe=royalblue(web), boxrule=2pt, arc=0mm, breakable]
{\scriptsize
\begin{lstlisting}[style=mystyle]
{
  "name": "generate_metrics",
  "args": {
    "metrics": [
      {
        "name": "Core Functionality and Web Stability",
        "description": "Evaluates whether the core functionality of the interface works correctly, the system is stable, and compatible across platforms.",
        "weight": 0.2,
        "criteria": [
          "Basic functionality working correctly",
          "System stability and error handling",
          "Cross-browser compatibility",
          "Mobile responsiveness"
        ]
      },
      ...
    ]
  }
}
\end{lstlisting}
}
\end{tcolorbox}

\input{figures/examples/2-stratege}
\input{figures/examples/3-iteration}
\input{figures/examples/6-visual_structure}
\input{figures/examples/7-interface_of_the_questionnaire}

\section{Supplementary Examples}
\label{appendix:visualizations}

\begin{itemize}
    \item Figure~\ref{fig:example_stratege} compares GenUI and ConvUI in \textit{Business Strategy \& Operations} task.
    \item Figure~\ref{fig:example_ci_iteration} shows the iterative refinement process for a continuous integration dashboard. Each version progressively enhances usability and clarity through structure-aware feedback.
    \item Figure~\ref{fig:visual_structure} demonstrates that the layout of GenUI significantly improves users’ perception of clarity, trustworthiness, and professionalism.
\end{itemize}

\section{Human Evaluation Questionnaire Interface}
\label{appendix: Human Evaluation Questionnaire Interface}

We show the annotation interfaces in Figure \ref{fig:7_interface_of_the_questionnaire_1}, \ref{fig:7_interface_of_the_questionnaire_2}, \ref{fig:7_interface_of_the_questionnaire_3}.

\section{Human Annotation Filtering}

To ensure the reliability of human annotations, we employed a multi-stage filtering process involving trap questions, consistency checks, and agreement rate evaluation.
\begin{itemize}\itemsep0em 
    \item \textbf{Trap Questions.} Each annotation task contained 8 UI comparison questions. In some questionnaires, we embedded trap questions in which the “UI” was not a real interface with components, but rather a simple instruction such as “Select Example A for all options” or “Select Example B for all options.” Annotators who failed to follow these explicit instructions were identified as inattentive, and their entire submissions were discarded.
    \item \textbf{Consistency Check.} We manually compared each annotator’s multiple-choice selections with their accompanying textual comments. If a comment stated that Example A was better but the selected option was B, we considered this a clear inconsistency indicative of random selection. Such annotations were removed.
    \item \textbf{Manual Review.} We conducted a manual review for annotators who had low agreement with other annotators and determined whether the annotator’s responses showed signs of random or careless selection. If so, all responses from that annotator were excluded.
\end{itemize}
Through this process, we ensured that the retained annotations were both attentive and internally consistent, thereby improving the overall quality of our evaluation.

\section{Real-User Query Evaluation}
\label{appendix: Real-User Query Evaluation}

We first collect user queries by presenting the following survey.

\begin{tcolorbox}[title={\textbf{\small  Query Collection Survey}}, colback=whitesmoke, colframe=royalblue(web), boxrule=2pt, arc=0mm, breakable]
{\small
Our study evaluates the quality and effectiveness of user interfaces for AI chatbots.

In the screening survey, you will be asked to tell us five different AI chatbot queries from your daily usage.

In the follow-up survey, you will be presented with pairs of interfaces in response to your queries and you need to carefully compare these interfaces and determine which one better addresses your query.

\textbf{Your Chatbot Queries}

Share 5 different \textbf{specific} queries you typically use ChatGPT, Claude, or other generative AI tools.

\textbf{These will be used to generate personalized interface comparisons for you to evaluate.}

For example:

What are some attractions in New York City?

How should I improve my tennis skills?

Explain artificial intelligence in a simple way.
}
\end{tcolorbox}

Note that the provided examples are randomly sampled from initial user submissions to ensure that users understand we are collecting specific queries, rather than high-level descriptions.
To ensure the quality of collected queries, we set a minimal number of characters required for each query to be 35. We specified the exact requirements for participants as the human annotation for the UIX benchmark, ensuring that each user is a regular user of ChatGPT/Claude or similar products.
Without any post-processing, all user queries are then used to generate one conversation interface and one generative interface. To allow for a high rate limit in this human study, we use Gemini-2.5-flash for both types of interfaces. These interfaces are then shown to the users (similar to Appendix Section \ref{appendix: Human Evaluation Questionnaire Interface}), and for each pair, we ask the user to provide an overall judgment of which kind of interface they prefer for their own query.

%% file: tables/2-llm_evaluation_results.tex
\begin{table*}[h]
\centering
\small
\renewcommand\arraystretch{1.2}
\setlength{\tabcolsep}{4pt}
\begin{tabular}{l|cc|ccc|cc}
\hline

\hline

\hline

\hline
\multirow{2}{*}{\textbf{Framework}} 
 & \multicolumn{2}{c|}{\textit{Functional}} 
 & \multicolumn{3}{c|}{\textit{Interactive}} 
 & \multicolumn{2}{c}{\textit{Emotional}} \\
& QIC & TaskEff 
& Usability & Learnability & IC 
& ASA & IES  \\
\hline
\multicolumn{3}{l}{\textit{- Score:}} \\
\hline
ConvUI (Claude 3.7) & 65.8 & 47.6 & 34.7 & 72.4 & 76.1 & 47.7 & 41.1 \\
ConvUI (GPT-4o)     & 70.2 & 51.0 & 36.8 & 74.9 & 80.2 & 48.1 & 43.1 \\
IUI                 & 68.0 & 58.0 & 57.9 & 73.8 & 72.5 & 70.8 & 56.0 \\
\rowcolor[gray]{.9}
GenUI               & \textbf{86.1} & \textbf{84.2} & \textbf{87.0} & \textbf{84.0} & \textbf{88.5} & \textbf{88.9} & \textbf{87.2} \\
\hline
\multicolumn{3}{l}{\textit{- Relative Improvement (\%):}} \\
\hline
\textit{vs.} ConvUI (Claude 3.7) 
& 30.9\% & 76.6\% & 151.0\% & 16.0\% & 16.2\% & 86.2\% & 112.4\% \\
\textit{vs.} ConvUI (GPT-4o) 
& 22.7\% & 65.1\% & 136.2\% & 12.2\% & 10.4\% & 84.8\% & 102.3\% \\
\textit{vs.} IUI 
& 26.7\% & 45.0\% & 50.2\% & 13.8\% & 22.0\% & 25.5\% & 55.7\% \\
\hline

\hline

\hline

\hline
\end{tabular}
\caption{
\textbf{LLM-Based Evaluation Scores Across Perception Dimensions.}
Automatic assessment (0–100 scale) of UI frameworks across functional, interactive, and emotional perception categories.
}
\label{tab:llm_evaluation_results}
\end{table*}

%% file: figures/examples/5-reward.tex
\begin{figure*}[t]
  \centering
  \begin{minipage}{\linewidth}
    \centering
    \textit{\textbf{User query:} “I want to understand quantum physics principles.”} \\
    \rule{0.95\linewidth}{0.4pt}
    \vspace{0.8em}
  \end{minipage}

  \setlength{\fboxsep}{0pt}
  \setlength{\fboxrule}{0.1pt}

  \begin{subfigure}[t]{0.4\linewidth}
    \centering
    \fcolorbox{black}{white}{\includegraphics[height=5.3cm]{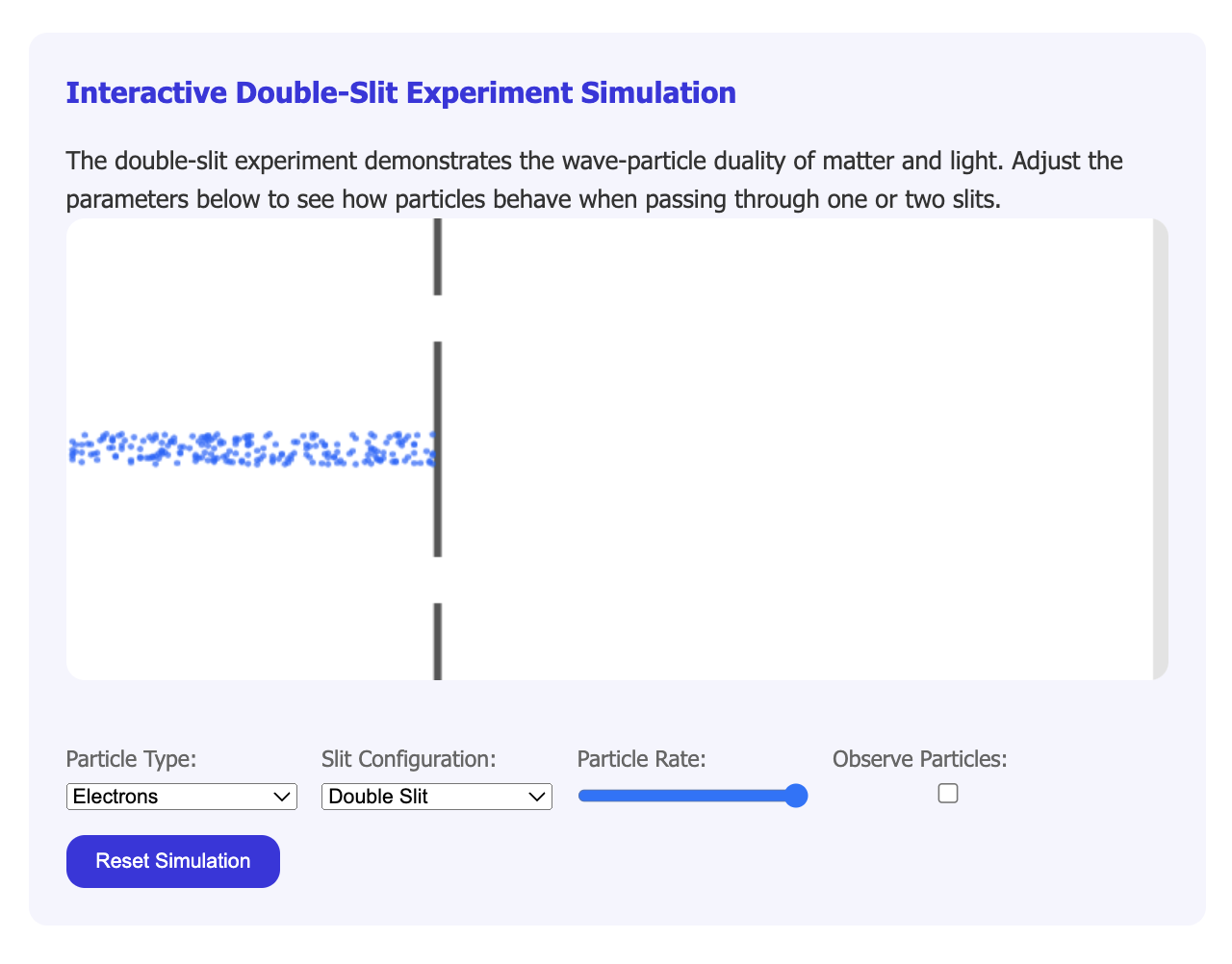}}
    \caption{\textbf{Static reward.} The simulation fails to visualize wave-particle duality.}
    \label{fig:static_reward}
  \end{subfigure}
  \hfill
  \begin{subfigure}[t]{0.58\linewidth}
    \centering
    \fcolorbox{black}{white}{\includegraphics[height=5.3cm]{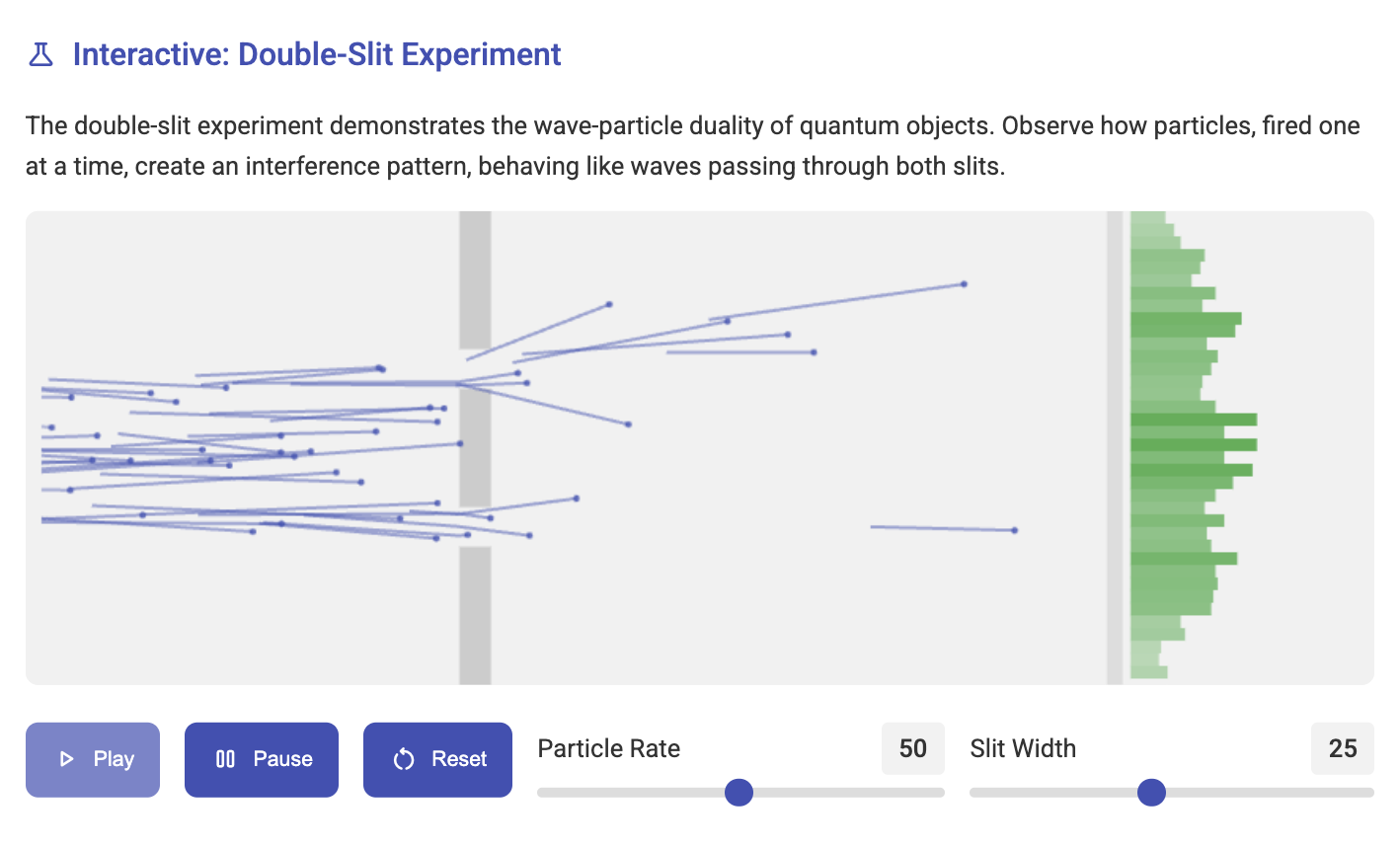}}
    \caption{\textbf{Dynamic reward.} The simulation successfully visualizes wave-particle duality.}
    \label{fig:dynamic_reward}
  \end{subfigure}

  \caption{\textbf{Visual comparison of static and dynamic reward settings.}}
  \label{fig:double_slit_visualization_comparison}
\end{figure*}

%% file: figures/examples/2-stratege.tex
\begin{figure*}[h]
  \centering
  \begin{minipage}{\linewidth}
    \centering
    \textit{\textbf{User query:} “I'm a consultant working with a family-owned manufacturing business that's been operating for 50 years. 
    They're facing increasing ... 
    How should they approach strategic transformation while preserving their heritage and retaining institutional knowledge?”} \\
    \rule{0.95\linewidth}{0.4pt}
    \vspace{0.8em}
  \end{minipage}

  \setlength{\fboxsep}{0pt}
  \setlength{\fboxrule}{0.1pt}

  \begin{subfigure}[t]{0.44\linewidth}
    \centering
    \fcolorbox{black}{white}{\includegraphics[height=5.6cm]{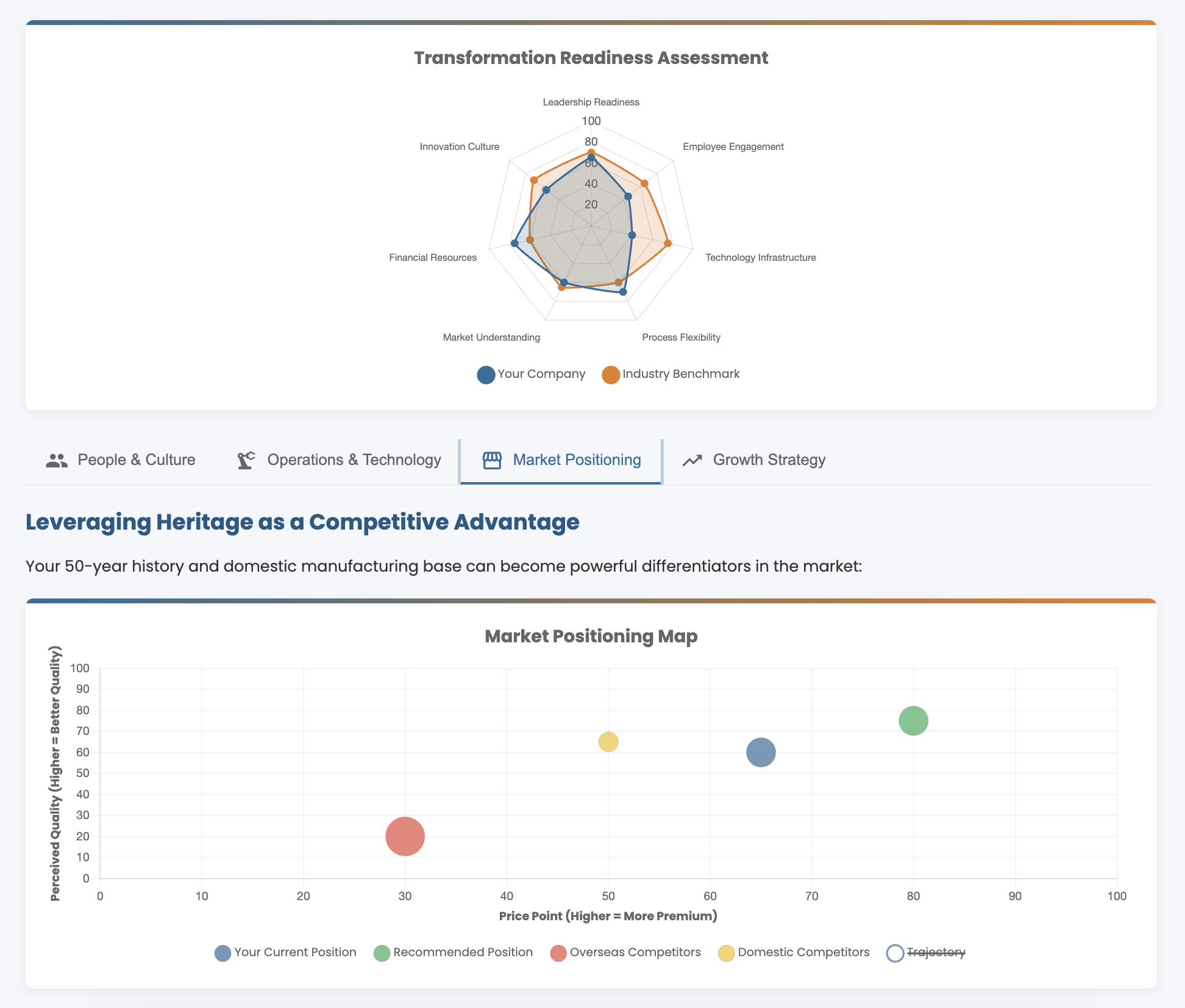}}
    \caption{\textbf{GeneUI} presents multiple charts and visual summaries.}
    \label{fig:genui_strategy}
  \end{subfigure}
  \hfill
  \begin{subfigure}[t]{0.54\linewidth}
    \centering
    \fcolorbox{black}{white}{\includegraphics[height=5.6cm]{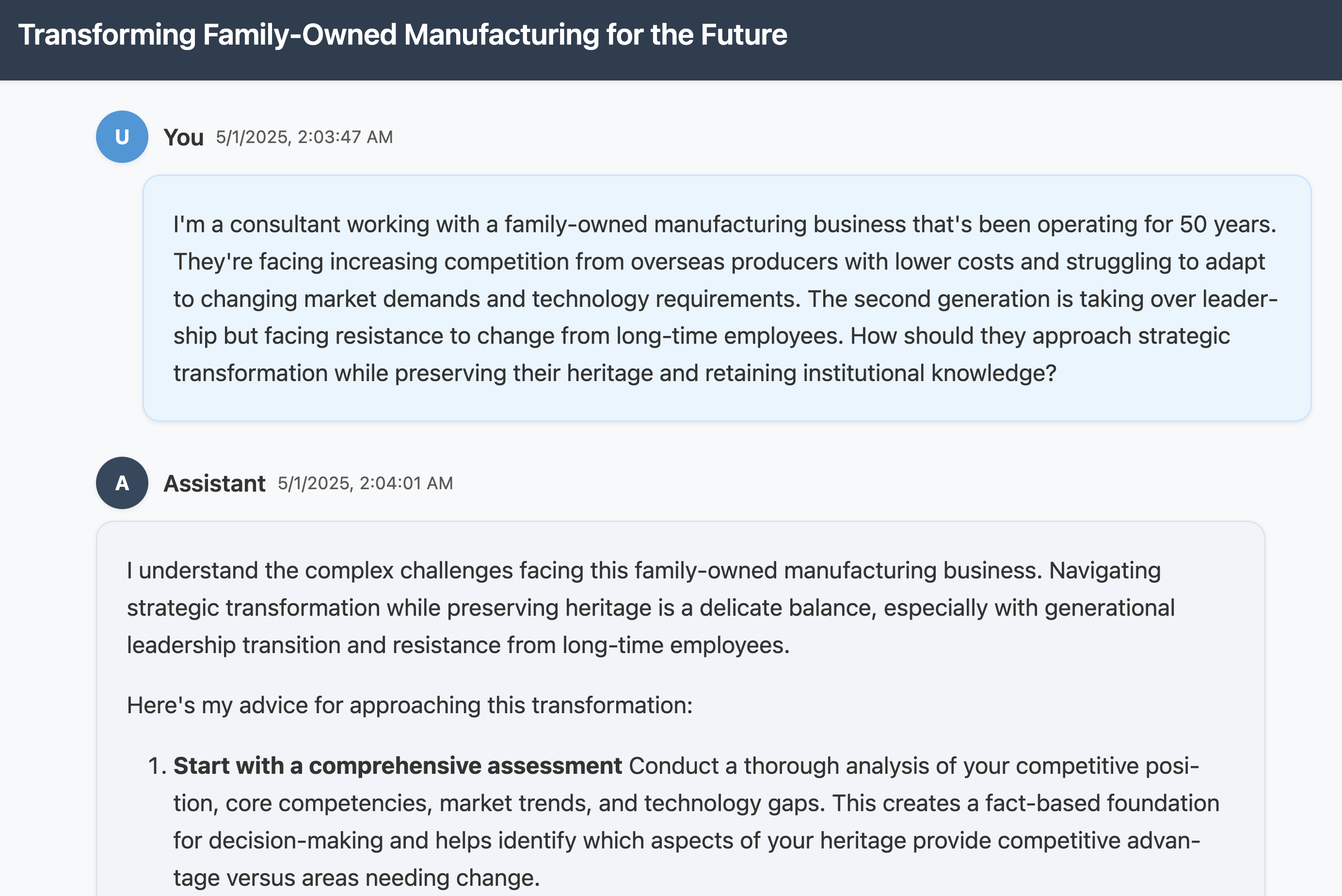}}
    \caption{\textbf{ConvUI} directly outlines strategy in a sectioned format.}
    \label{fig:convui_strategy}
  \end{subfigure}

  \caption{\textbf{GenUI \textit{vs.} ConvUI} in \textit{Business Strategy \& Operations} task.}
  \label{fig:example_stratege}
\end{figure*}

%% file: figures/examples/3-iteration.tex
\begin{figure*}[h]
  \centering
  \begin{minipage}{\linewidth}
    \centering
    \textit{\textbf{User query:} “I want to set up a continuous integration workflow.”} \\
    \rule{0.95\linewidth}{0.4pt}
    \vspace{0.8em}
  \end{minipage}

  \setlength{\fboxsep}{0pt}
  \setlength{\fboxrule}{0.1pt}

  \begin{subfigure}[t]{0.48\linewidth}
    \centering
    \fcolorbox{black}{white}{\includegraphics[height=5cm]{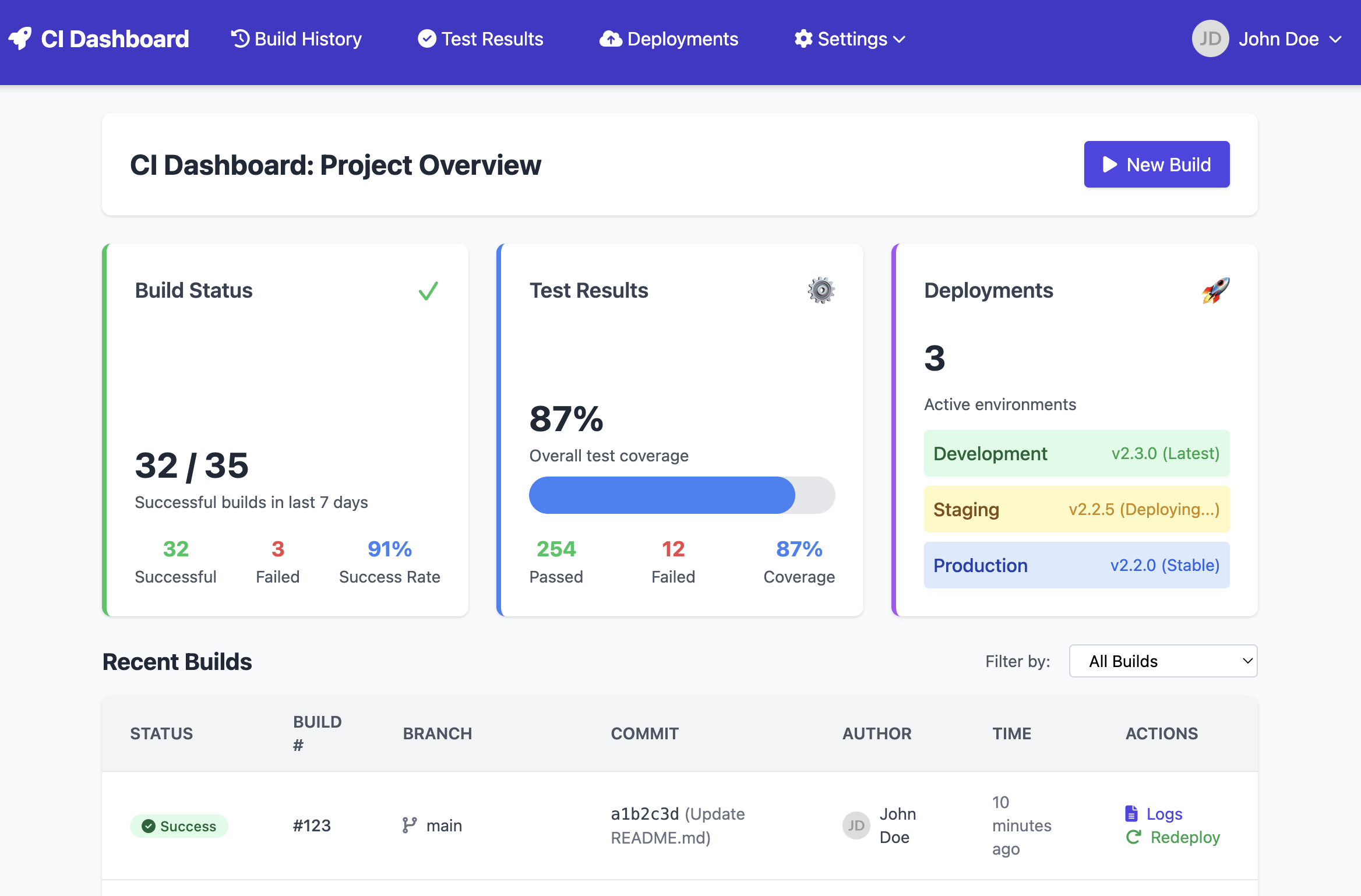}}
    \caption{\textbf{Iteration 1:} A basic CI dashboard with textual build/test summaries and limited interaction affordances.}
    \label{fig:iteration_1}
  \end{subfigure}
  \hfill
  \begin{subfigure}[t]{0.48\linewidth}
    \centering
    \fcolorbox{black}{white}{\includegraphics[height=5cm]{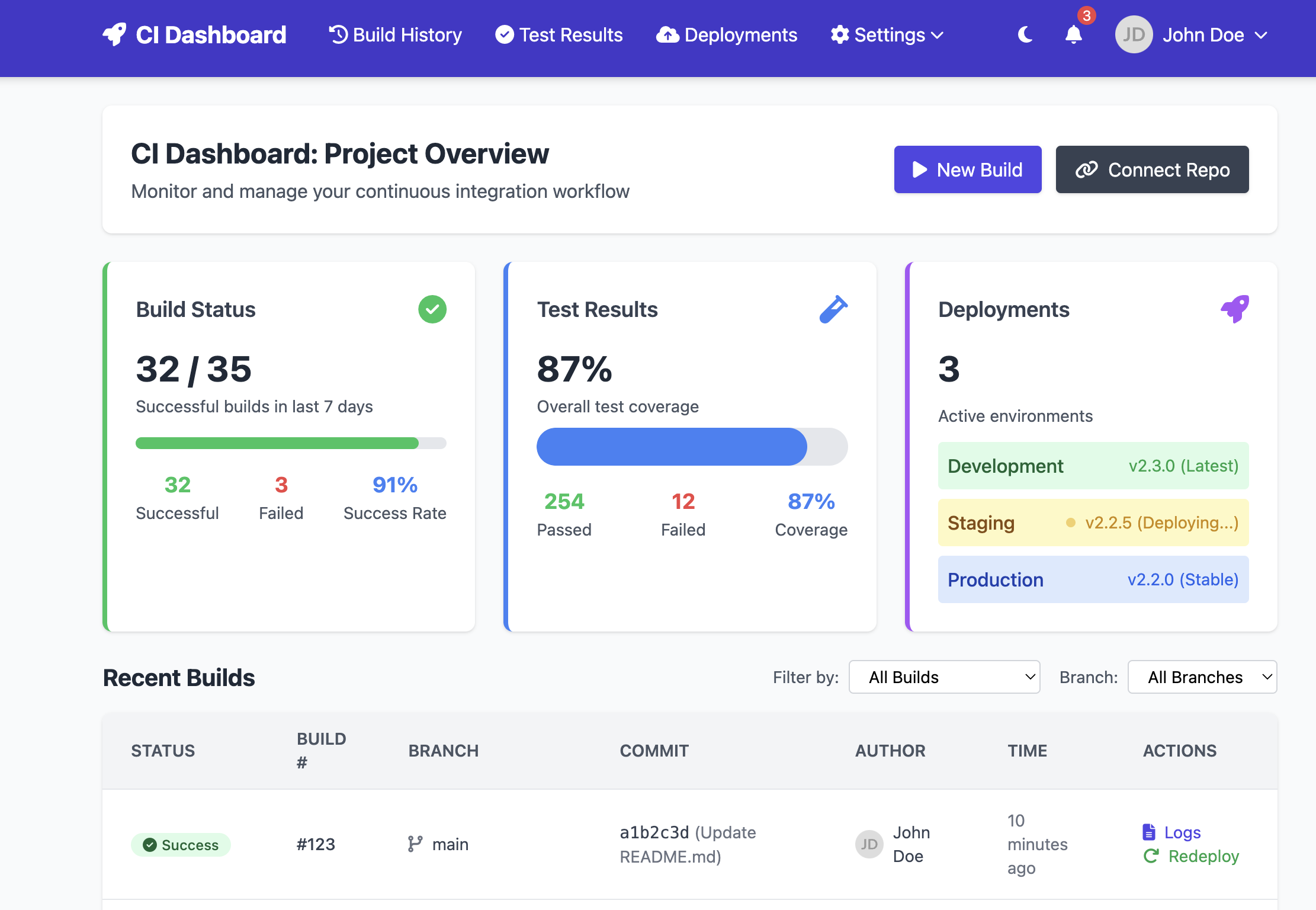}}
    \caption{\textbf{Iteration 2:} Improves layout compactness by closing excessive gaps and clarifies the CI context with stronger visual grouping.}
    \label{fig:iteration_2}
  \end{subfigure}
  \vspace{0.8em}

  \begin{subfigure}[t]{0.48\linewidth}
    \centering
    \fcolorbox{black}{white}{\includegraphics[height=5.5cm,width=7.6cm]{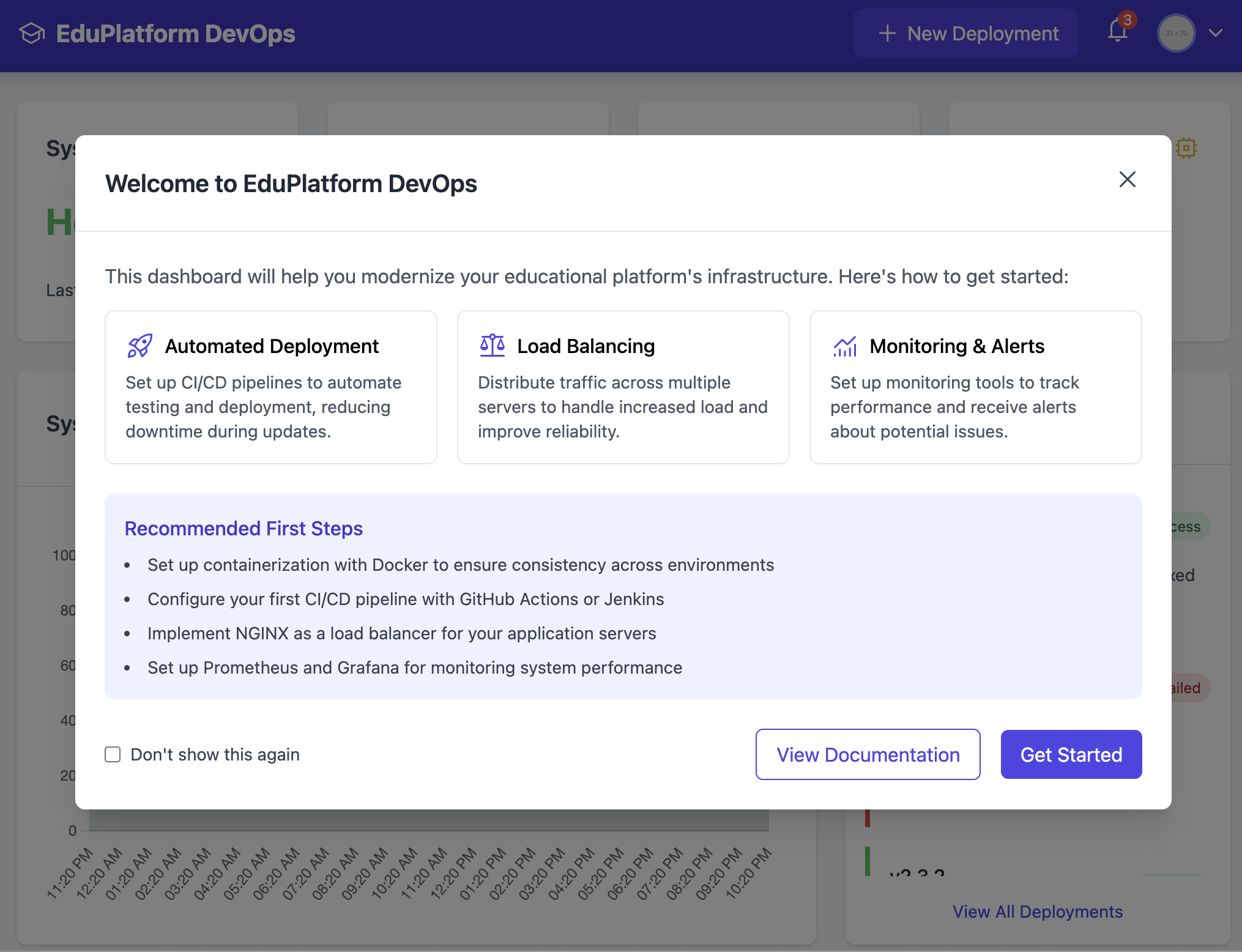}}
    \caption{\textbf{Iteration 3 (Onboarding page):} Introduces an onboarding modal outlining key components and recommended first steps.}
    \label{fig:iteration_3a}
  \end{subfigure}
  \hfill
  \begin{subfigure}[t]{0.48\linewidth}
    \centering
    \fcolorbox{black}{white}{\includegraphics[height=5.5cm]{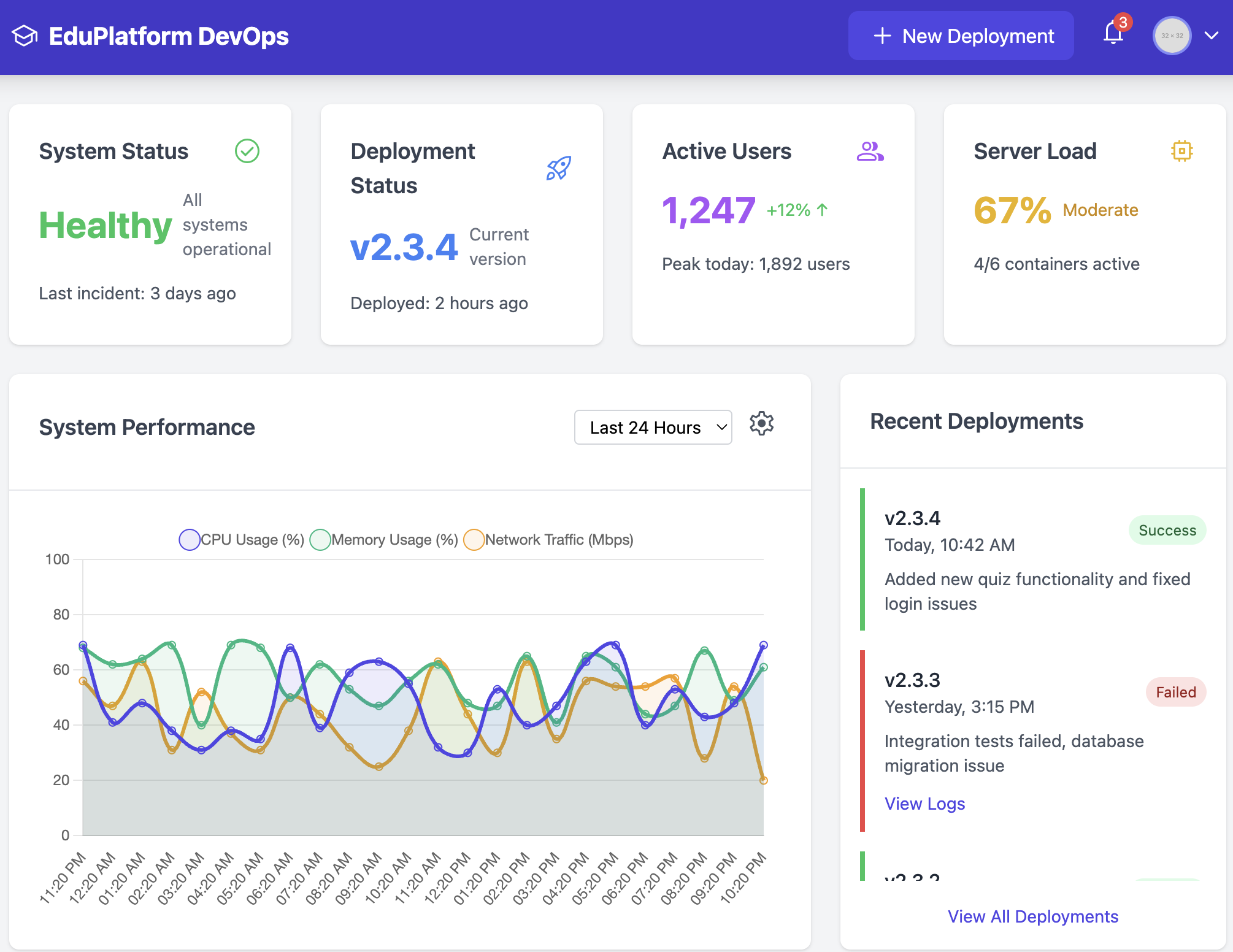}}
    \caption{\textbf{Iteration 3 (Main page):} Refactors layout to present deployment insights visually, using charts to highlight system status and activity trends.}
    \label{fig:iteration_3b}
  \end{subfigure}

  \caption{\textbf{Evolution across UI iterations} for the \textit{Continuous Integration Workflow} setup. Each version builds upon its predecessor by reducing visual clutter, providing onboarding guidance, and progressively enhancing the clarity of system performance and CI process feedback.}
  \label{fig:example_ci_iteration}
\end{figure*}

%% file: figures/examples/6-visual_structure.tex
\begin{figure*}[h]
  \centering
  \begin{minipage}{\linewidth}
    \centering
    \textit{\textbf{User query:} “How do I conduct market research?”} \\
    \rule{0.95\linewidth}{0.4pt}
    \vspace{0.8em}
  \end{minipage}

  \setlength{\fboxsep}{0pt}
  \setlength{\fboxrule}{0.1pt}

  \begin{subfigure}[t]{0.49\linewidth}
    \centering
    \fcolorbox{black}{white}{\includegraphics[height=5cm]{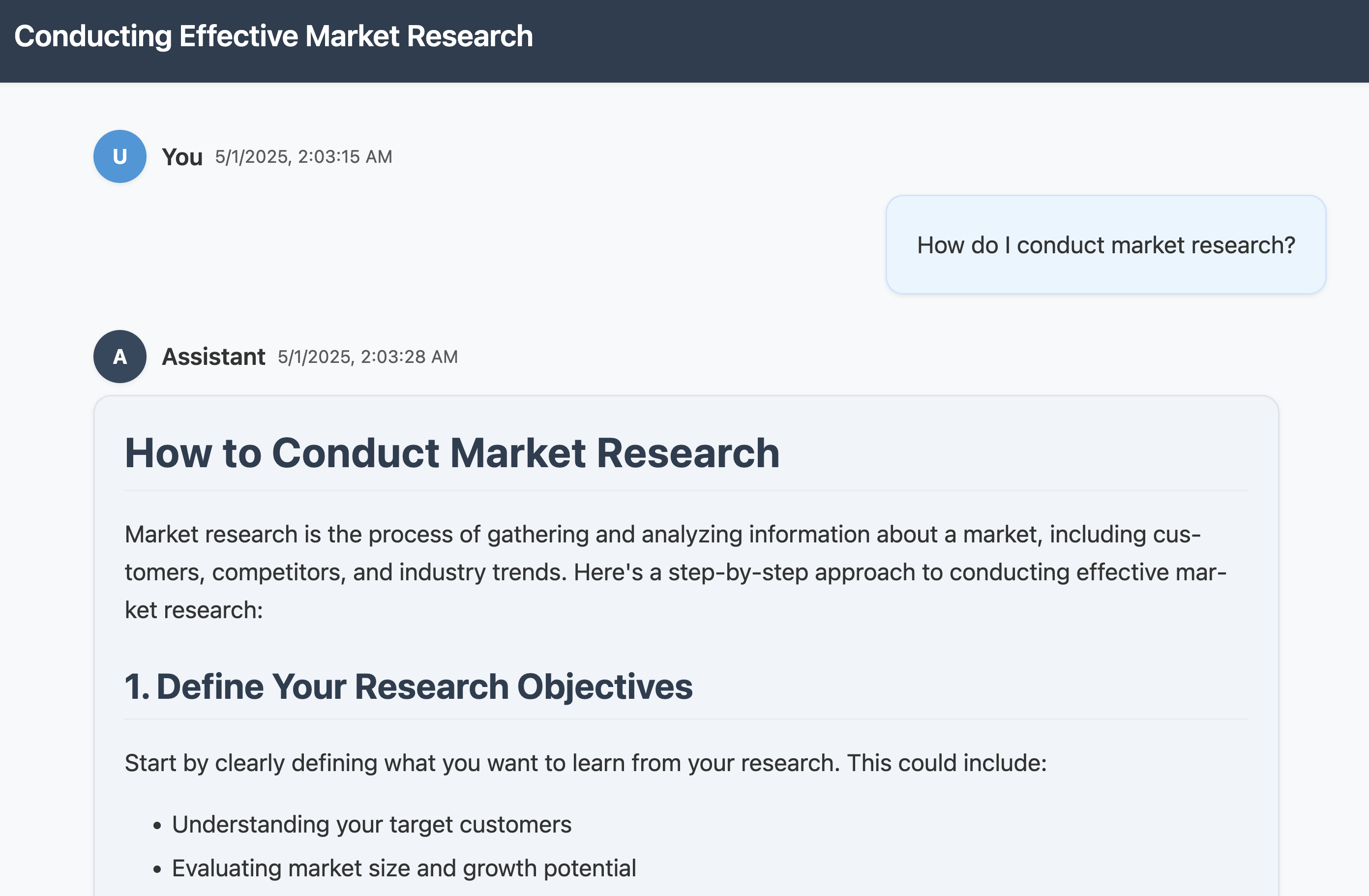}}
    \caption{\textbf{ConvUI.} Presents information as plain linear text without visual hierarchy, making it harder to navigate.}
    \label{fig:visual_structure_convui}
  \end{subfigure}
  \hfill
  \begin{subfigure}[t]{0.49\linewidth}
    \centering
    \fcolorbox{black}{white}{\includegraphics[height=5cm]{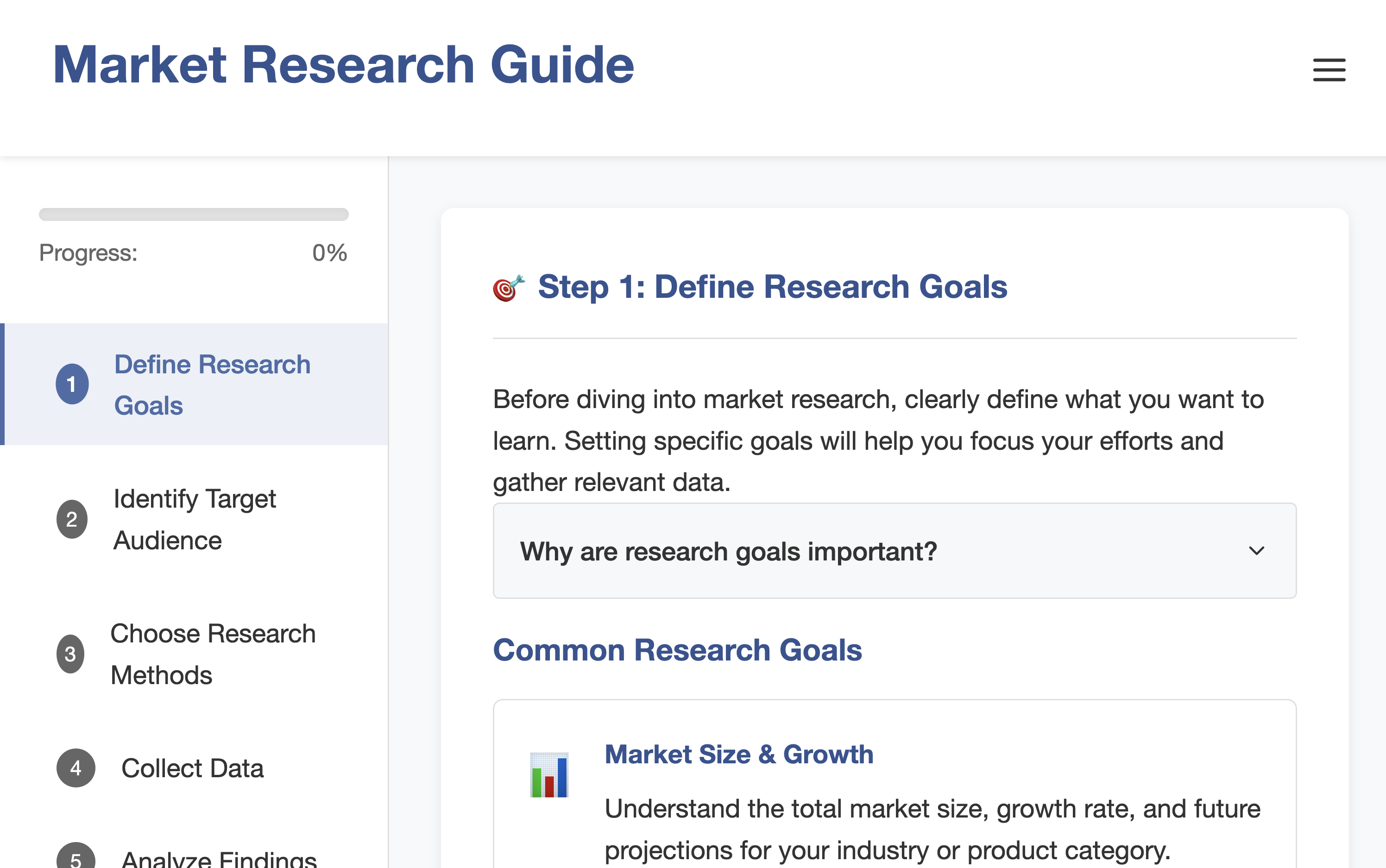}}
    \caption{\textbf{GenUI.} Organizes content into modular sections with clear structure, guiding users through the research process.}
    \label{fig:visual_structure_genui}
  \end{subfigure}

  \caption{\textbf{Visual structure enhances perceived professionalism.} Despite conveying similar content, GenUI was consistently rated as more trustworthy and well-organized due to its structured layout and visual clarity.}
  \label{fig:visual_structure}
\end{figure*}

%% file: figures/examples/7-interface_of_the_questionnaire.tex
\begin{figure*}[h]
  \centering
\includegraphics[width=0.8\linewidth]{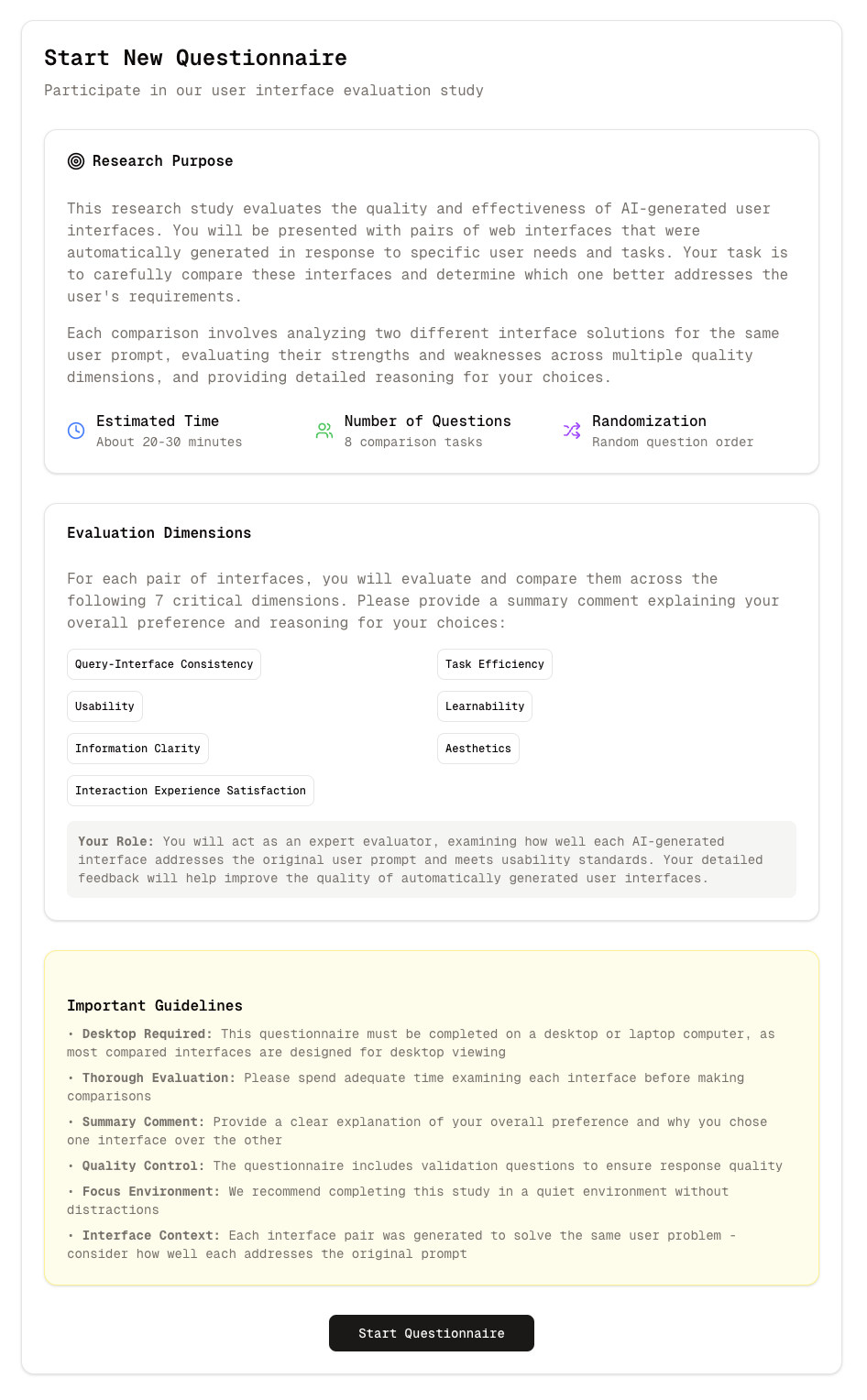}
  \caption{Human Evaluation Questionnaire Interface (a)
  }
  \label{fig:7_interface_of_the_questionnaire_1}
\end{figure*}

\begin{figure*}[h]
  \centering
\includegraphics[width=0.8\linewidth]{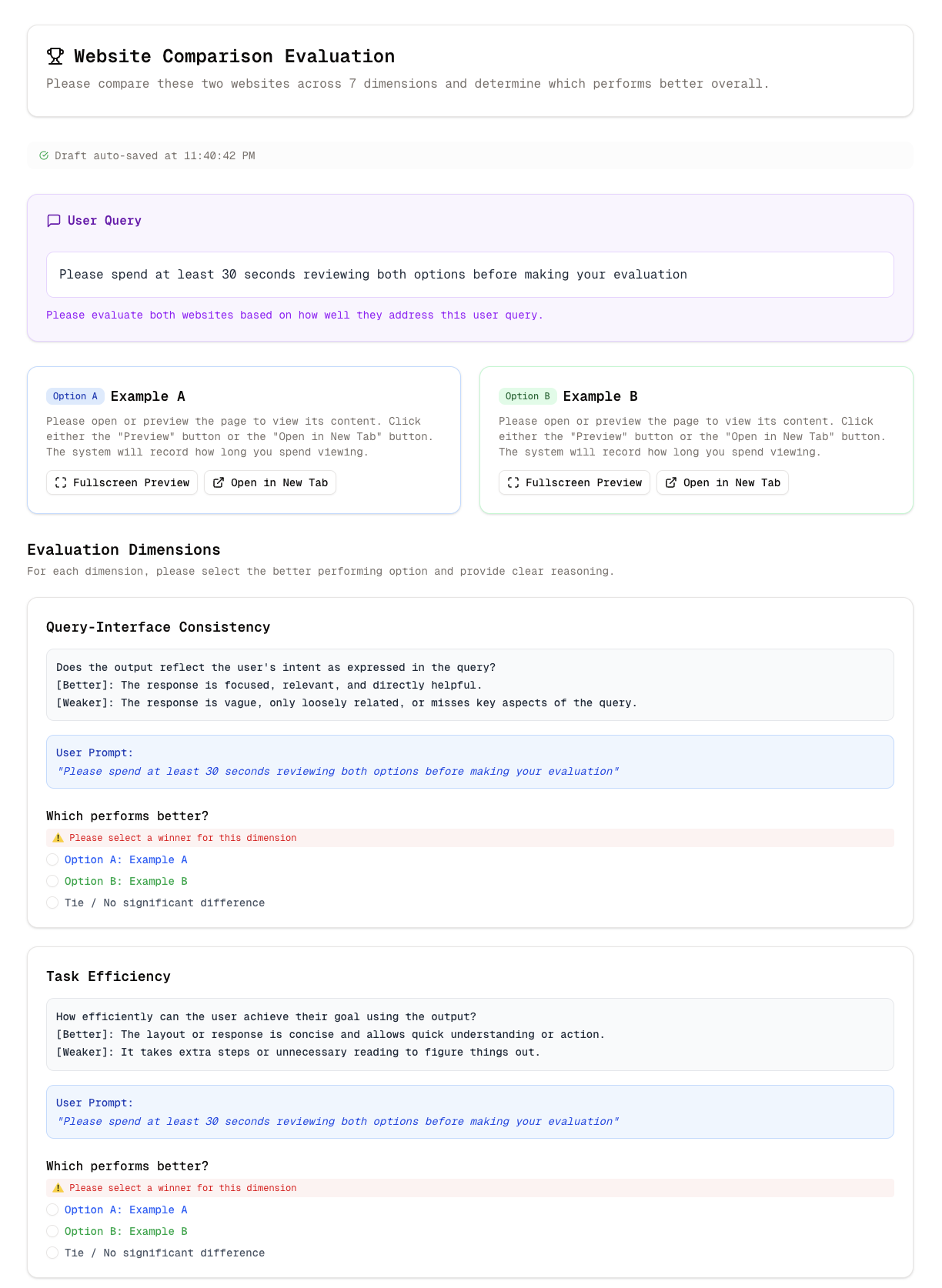}
  \caption{Human Evaluation Questionnaire Interface (b)
  }
  \label{fig:7_interface_of_the_questionnaire_2}
\end{figure*}

\begin{figure*}[h]
  \centering
\includegraphics[width=0.8\linewidth]{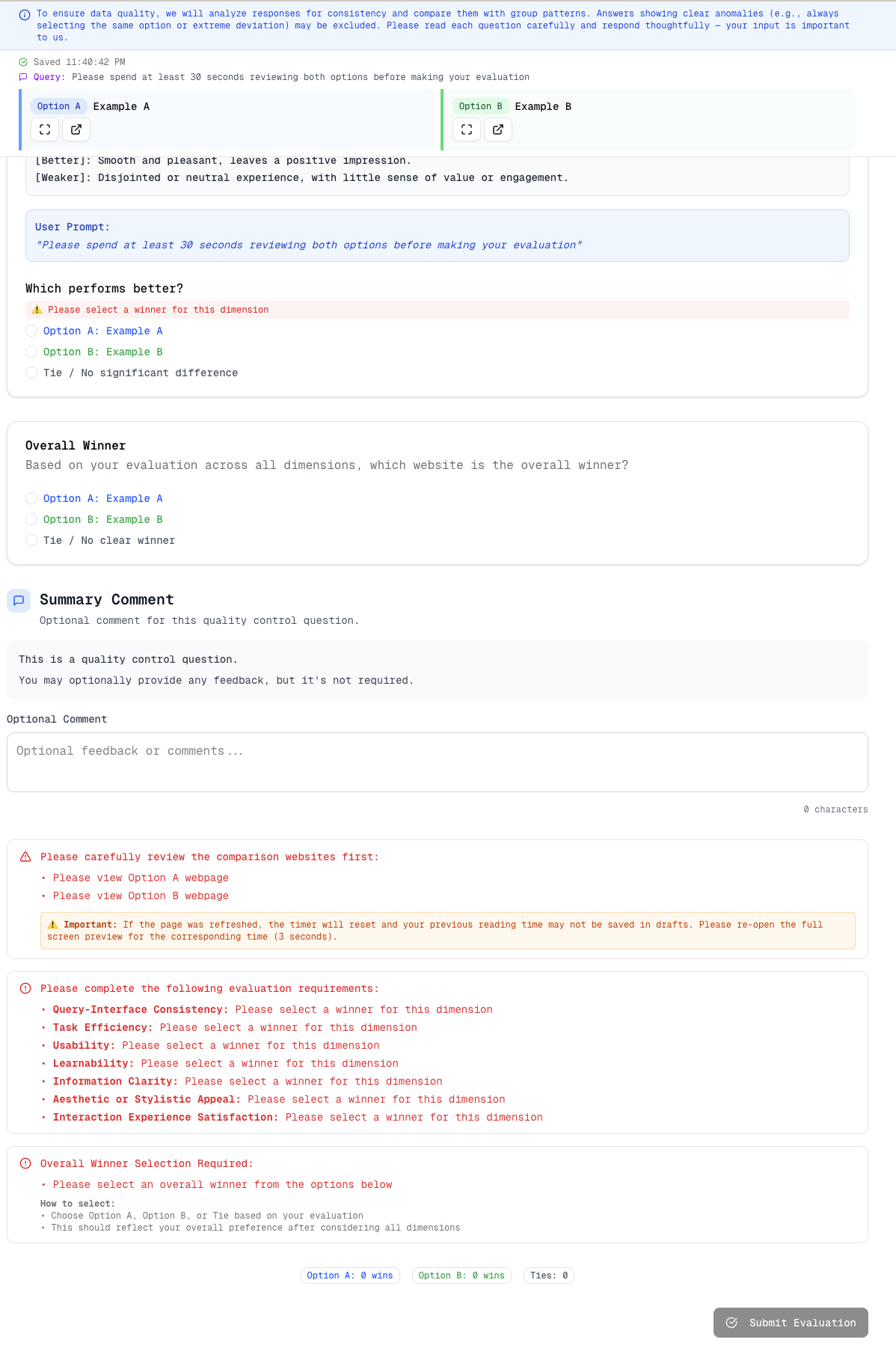}
  \caption{Human Evaluation Questionnaire Interface (c)
  }
  \label{fig:7_interface_of_the_questionnaire_3}
\end{figure*}